\def\ASUKAWithSupplement{}
\documentclass[lettersize,journal]{IEEEtran}
\usepackage{amsmath,amsfonts}
\usepackage{algorithmic}
\usepackage{algorithm}
\usepackage{array}
\usepackage[caption=false,font=normalsize,labelfont=sf,textfont=sf]{subfig}
\usepackage{textcomp}
\usepackage{stfloats}
\usepackage{url}
\usepackage{verbatim}
\usepackage{graphicx}
\usepackage{cite}
\usepackage[normalem]{ulem}
\usepackage{hyperref}
\usepackage{enumitem}
\usepackage{bm}
\usepackage{booktabs}
\usepackage{xcolor}
\usepackage{threeparttable}
\renewcommand{\paragraph}[1]{\noindent\textbf{#1}\ \ }
\usepackage{colortbl}
\definecolor{rank1}{RGB}{226, 164, 145} 
\definecolor{rank2}{RGB}{235, 197, 185} 
\definecolor{rank3}{RGB}{244, 227, 222} 
\usepackage{multirow}
\usepackage{makecell}
\usepackage{xcolor}

\begin{document}

\title{Aligned Stable Inpainting: Mitigating Unwanted Object Insertion and Preserving Color Consistency}

\author{
Yikai Wang*, Junqiu Yu*, Chenjie Cao, Xiangyang Xue, Yanwei Fu
\thanks{
Yikai and Junqiu contribute equally. Yanwei Fu is the corresponding author.}
\thanks{Yikai Wang is with Nanyang Technological University. 
Chenjie Cao is with Tencent Hunyuan3D.
Junqiu Yu, Xiangyang Xue, Yanwei Fu are with Fudan University, Shanghai 200437, China. 
Yanwei Fu is also with the Fudan ISTBI–ZJNU Algorithm Centre for Brain-Inspired Intelligence, Zhejiang Normal University, Jinhua 321017, China, and also with Shanghai Innovation Institute, Shanghai 200080, China.
}
\thanks{Project page: \url{https://yikai-wang.github.io/asuka}}
\thanks{
E-mails: yi-kai.wang@outlook.com; yanweifu@fudan.edu.cn.
}
}

\markboth{Journal of \LaTeX\ Class Files,~Vol.~14, No.~8, August~2021}%
{Shell \MakeLowercase{\textit{et al.}}: A Sample Article Using IEEEtran.cls for IEEE Journals}

\IEEEpubid{0000--0000/00\$00.00~\copyright~2021 IEEE}

\maketitle

\begin{abstract}
Generative image inpainting can produce realistic results even with large, irregular masks, but existing methods still suffer from two common problems:
(1) Unwanted object insertion: 
hallucinate artifacts that do not match the surrounding context.
(2) Color inconsistency: 
noticeable color shifts that lead to smeared textures.
We analyze the causes of these issues and propose Aligned Stable inpainting with UnKnown Areas prior (ASUKA), a post-hoc framework for pre-trained inpainting models.
To reduce unwanted object insertion, we use reconstruction-based priors to guide the generative model, suppressing hallucinated objects while preserving generative flexibility.
To address color inconsistency, we design a specialized VAE decoder that formulates latent-to-image decoding as a local harmonization task. 
We implement ASUKA on both U-Net-based and DiT-based inpainting models with lightweight modifications.
Experiments on Places2 and MISATO, our proposed benchmark, show that ASUKA effectively suppresses object hallucination and improves color consistency, outperforming existing diffusion- and rectified flow-based inpainting methods. 
The dataset, models, and code will be released on GitHub.
\end{abstract}


\section{Introduction}
\begin{figure}
\centering
\includegraphics[width=0.9\linewidth]{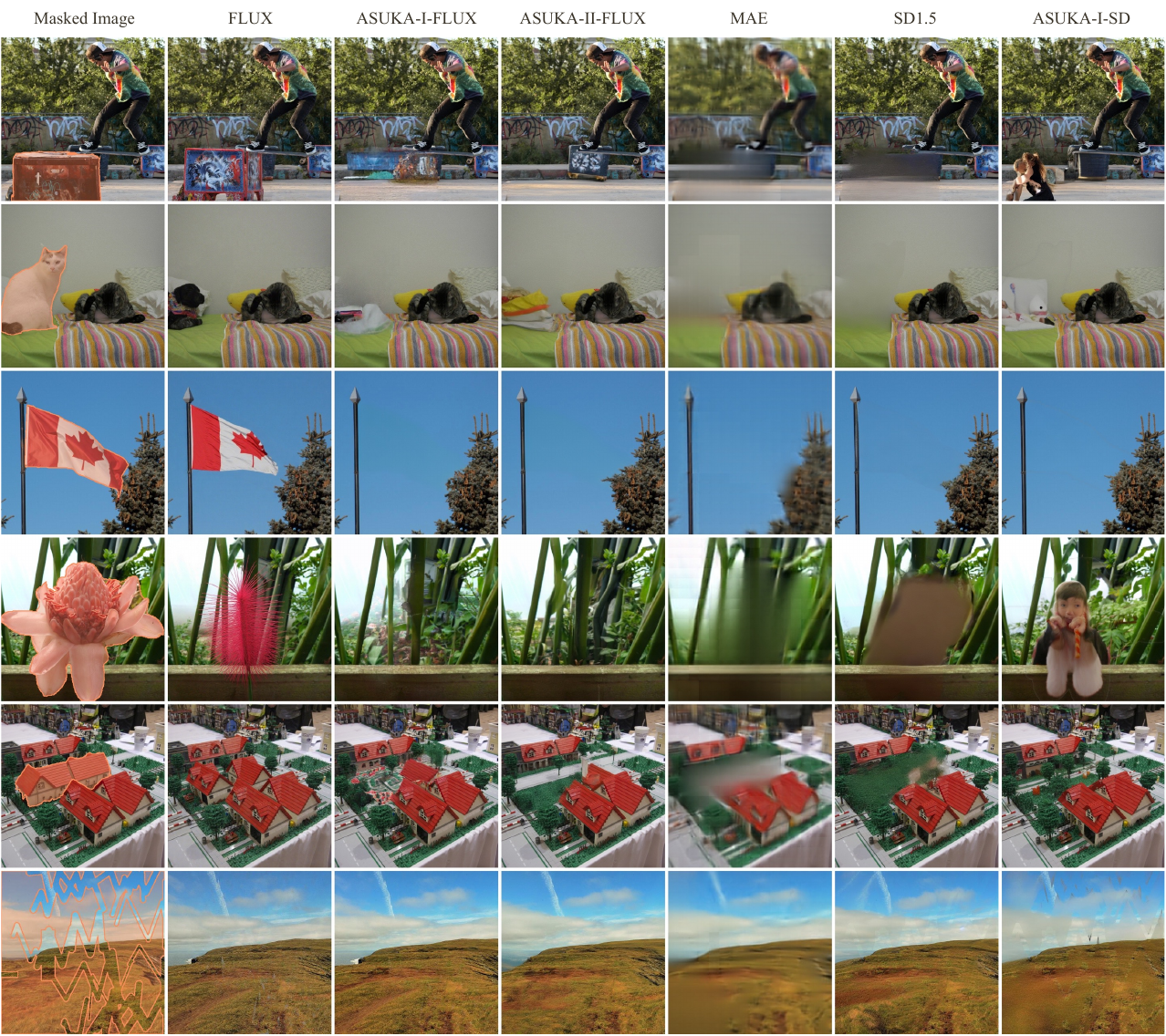}
\caption{
Comparison on image inpainting results.
Existing latent inpainting methods suffer from two major issues:
unwanted object insertion and color inconsistency.
ASUKA introduces a post-training procedure to address these two issues.
ASUKA-II suppresses unwanted object insertion more effectively and further improves color consistency compared with ASUKA-I.
}
\label{fig:object_removal}
\end{figure}
Image inpainting~\cite{bertalmio2000image} restores missing parts of an image while keeping them consistent with the visible regions.
Traditional methods~\cite{bertalmio2000image,Criminisi2003ObjectRB,hays2007scene,Levin2003LearningHT,Roth2005FieldsOE} often produce blurry results when reconstructing masked areas~\cite{pathak2016context}.
Models based on Generative Adversarial Networks (GANs) can handle complex mask shapes and have achieved impressive inpainting results~\cite{nazeri2019edgeconnect,liao2020guidance,cao2021learning,suvorov2022resolution,wan2021highfidelity,zhao2021large,goodfellow2014generative,ho2020denoising,esser2021taming}.
However, they still face difficulties in challenging scenarios, especially when filling large missing regions.
More recently, powerful generative models such as Stable Diffusion~\cite{Rombach_2022_CVPR} and FLUX~\cite{flux} have emerged as versatile tools for inpainting, thanks to their large capacity and extensive training datasets.
These models typically follow a latent-generation pipeline: the image is first encoded into a compact latent space, and the inpainting model is then trained within this space.

However, latent-based generative inpainting models still face some challenges, which make the inpainted results less faithful to the original image. In particular:

\IEEEpubidadjcol
\noindent(1) \textit{Unwanted object insertion}: The model sometimes generates random, irrelevant elements in the masked regions (see the first and fourth rows of Fig.~\ref{fig:object_removal}).
We attribute this issue to three factors. 
First, random masking with global captions teaches the model to reconstruct masked objects even when they are not mentioned. 
Second, residual cues in the visible region, such as shadows or nearby objects, can trigger object regeneration. 
Third, attention-based generators may rely too much on their learned image prior instead of user or context guidance. 
While carefully adjusting prompts can reduce this issue, the best choice of prompt depends on the specific image, making this solution impractical in real-world applications.

\noindent(2) \textit{Color inconsistency}:
This issue, though less studied in academia but critical in practice, leads to noticeable differences in color between the inpainted and original regions. These include mismatches in brightness, saturation, luminance, and hue, causing smear-like artifacts, as shown in the second and last rows of Fig.~\ref{fig:object_removal}.
The root cause is a mismatch between the pixel distributions of the filled regions and the original image, due to limitations in the latent generative model and VAE (see Fig.~\ref{fig:color-shift}). 
For image generation, this problem is less severe because the whole image is generated and the color shift remains consistent across generated pixels. 
However, for inpainting, the inconsistency is significant since the unmasked regions have ground-truth pixels. 

To reduce object hallucination and improve color consistency in image inpainting, we introduce the ASUKA (Aligned Stable inpainting with UnKnown Areas prior) framework, improving latent inpainting models by combining regression-based reconstruction with distribution-aligned generation.
Specifically, we refine the generation and decoding stages to limit object hallucination and enhance color consistency.
(1) \emph{To mitigate object hallucination},
we use the Masked Auto-Encoder (MAE)~\cite{he2022masked} as a prior to replace the text condition to guide and stabilize the generation process.
As shown in Fig.~\ref{fig:object_removal}, MAE produces stable but slightly blurred outputs, whereas generative models can create unrealistic content despite their strong generation ability.
By combining the MAE prior with latent generative models, we reduce object hallucination while maintaining performance.
(2)
\emph{To handle color mismatches} between masked and unmasked regions, we redesign the VAE decoder to act as a local harmonization module conditioned on the visible pixels.
\uline{Notably, our decoder can serve as a plug-and-play component to enhance general inpainting models, including text-guided inpainting.}

In our conference paper~\cite{wang2025towards} (CVPR 2025 highlight paper), the preliminary version of ASUKA, ASUKA-I, was initially designed to U-Net based SD1.5 inpainting model, which we denote as ASUKA-I-SD.
As transformer backbones have become increasingly popular in image generation tasks, in this manuscript, we first adapt ASUKA-I to a pure transformer model, FLUX.1-Fill-dev~\cite{flux}, leading to ASUKA-I-FLUX.
The main findings of ASUKA-I-FLUX are consistent with those of ASUKA-I-SD, but several issues arise that limit its generalization:
(1) \emph{Complex text conditioning}: 
In MMDiT-style models, the text condition serves as the model input rather than a frozen cross-attention condition.
This deep transformation of the text condition makes it difficult to align with MAE features, leading to unstable MAE conditioning.
(2) \emph{Position modeling}: Transformer-based models require explicit modeling of input condition positions. When such information is missing, the model’s controllability is reduced.
(3) \emph{Limited decoder generalization}: The fine-tuned decoder struggles to generalize to realistic masks in real-world applications.

To address these issues, we systematically redesign the ASUKA framework and propose ASUKA-II, which further reduces unwanted object insertion and better preserves color consistency.
Inspired by the success of ASUKA-I-SD, we introduce a cross-attention–based control module across all cross-attention layers. 
This design uses a shared MAE feature combined with layer-specific shallow adapters to better inject MAE information.
This approach avoids transforming MAE features through the complex text-conditioning pathway, thereby improving control stability.
We also employ scaled positional encoding to ensure that the fixed low-dimensional MAE features can guide high-resolution image generation.
Finally, we enhance the decoder’s training recipe to improve its generalization performance in real-world applications.
These steps allow the ASUKA series to reduce object hallucination and produce more color-consistent inpainting results.

To assess performance under different scenarios and mask shapes, we go beyond the standard Places2 dataset~\cite{zhou2017places} and introduce an evaluation dataset called MISATO. MISATO is built from \underline{M}atterport3D~\cite{Matterport3D}, Fl\underline{i}ckr-Land\underline{s}cape~\cite{lin2021infinitygan}, Meg\underline{a}Dep\underline{t}h~\cite{MDLi18}, and C\underline{O}CO 2014~\cite{lin2014microsoft},
spanning four distinct domains including landscape, indoor, building, and background, providing diverse coverage for benchmarking.
Experiments on both MISATO and Places2 with large irregular masks confirm the effectiveness of ASUKA.

\paragraph{Contributions} 
ASUKA improves image inpainting by keeping colors consistent and reducing object hallucination, while preserving the generation ability of a frozen inpainting model. It does this with two main components:
(1) \emph{Context-stable alignment}: ASUKA aligns the stable MAE prior with generative models to give a reliable estimate of masked regions, using the MAE prior instead of text prompts.
(2) \emph{Color-consistent alignment}: ASUKA treats the decoding process from latent space to image as a local harmonization task. It trains a decoder specialized for inpainting to better blend masked and unmasked regions, which reduces color mismatches.

\paragraph{Extensions}
In our conference version~\cite{wang2025towards}, we introduced ASUKA-I and applied it to a U-Net-style inpainting model, SD1.5.
In this manuscript, we provide a thorough analysis of the above mentioned two issues.
Based on these analysis, we first extend ASUKA-I to a multi-modal transformer-based inpainting model, FLUX.1-Fill-dev.
We then analyze and identify several issues that arise when the controlling mechanisms and model scale change.
To address these issues, we propose ASUKA-II, which further improves the system’s ability to prevent unwanted object insertion and maintain color consistency.
Specifically:
(1) We introduce a cross-attention-based control injection module to avoid fitting the complex text-conditioning component.
(2) We adopt a scaled positional encoding mechanism to enhance control performance for high-resolution images using a fixed low-resolution MAE prior.
(3) We refine the decoder training strategy to improve color consistency in real-world applications.

\section{Related Works}

\paragraph{Image inpainting} fills in missing parts of an image with realistic content.
Traditional methods, such as patch matching~\cite{criminisi2004region,barnes2009patchmatch,zhang2018robust} or differential equation approaches~\cite{bertalmio2000image,chan2001nontexture,bertalmio2003simultaneous}, rely on low-level features and often fail when dealing with large gaps.
Deep context-aware methods learn stronger priors from data: CNN- and GAN-based approaches introduce mask-aware convolutions and normalization~\cite{liu2018image,yu2019free}, contextual attention~\cite{yu2018generative,yi2020contextual,zeng2020high,ko2023continuously}, adversarial learning~\cite{pathak2016context,yu2019free,zhao2021large,zeng2022aggregated,sargsyan2023mi}, and frequency-aware modeling~\cite{suvorov2022resolution,xu2023image,chu2023rethinking}.
Recent methods such as MAT~\cite{li2022mat}, MAE-FAR~\cite{cao2022learning}, EdgeConnect~\cite{nazeri2019edgeconnect}, ZITS/ZITS++~\cite{dong2022incremental, cao2023zits++} further improve long-range reasoning by leveraging attention, masked reconstruction, edges, structural tokens, or Fourier priors, but reconstruction-driven objectives can still produce blurry or overly conservative results for large masks. 
More recently, diffusion and rectified-flow models, including Palette~\cite{saharia2022palette}, RePaint~\cite{lugmayr2022repaint}, SD~\cite{Rombach_2022_CVPR}, Kandinsky~\cite{arkhipkin2023kandinsky}, TurboFill~\cite{xie2025turbofill}, and large-scale flow-based editors~\cite{esser2024scaling,flux} and most recently works~\cite{lu2025pinco, vu2026inverfill, chow2026editmgt, zheng20262}, have shown stronger generative realism, while specialized methods such as PixelHacker~\cite{xu2025pixelhacker}, OmniPaint~\cite{yu2025omnipaint}, OmniEraser~\cite{wei2025omnieraser}, BrushNet~\cite{ju2024brushnet}, PowerPaint~\cite{zhuang2024task}, HD-Painter~\cite{manukyan2025hd} explore semantic priors, object-oriented removal/insertion, adapter-based editing, and high-resolution completion. 
Despite these advances, robust context-aware completion remains challenging.

\paragraph{Adapting latent generative models} 
Latent diffusion models (LDMs)~\cite{Rombach_2022_CVPR} and rectified flow models~\cite{esser2024scaling, flux} 
combine a VAE, which learns a compact latent space, with a generative model that operates in this latent space to accelerate generation.
Many approaches have been proposed to add new conditions to these models, including image inversion for text-guided image translation~\cite{meng2021sdedit}, textual inversion for personalization~\cite{gal2022image}, LoRA fine-tuning~\cite{hu2021lora}, and ControlNet~\cite{zhang2023adding} for incorporating diverse types of guidance~\cite{wang2024repositioning,cao2024leftrefill, gong2026direct}.
For inpainting, we drop the text condition and instead guide generation via a Masked Auto-Encoder (MAE)~\cite{he2022masked} prior.

\paragraph{Information loss in latent inpainting models}
VAE used in latent generative models often introduces distortions when reconstructing images.
In addition, we empirically found that the mismatch between generated latents and real latents leads to color inconsistencies (see Fig.~\ref{fig:color-shift} and Fig.~\ref{fig:latent-aug-visualization} for examples).
OpenAI~\cite{openai-decoder} uses a larger decoder to improve the decoding quality of Stable Diffusion’s latents.
Luo \emph{et al.}~\cite{luo2023image} introduce a frequency-augmented decoder for the super-resolution setting.
Zhu \emph{et al.}~\cite{zhu2023designing} propose preserving unmasked regions during decoding.
In this work, we focus on maintaining color consistency throughout the decoding process.

\paragraph{Masked Image-Modeling~\cite{bao2021beit}} (MIM) is a central topic in self-supervised learning.
Typical MIM methods~\cite{bao2021beit,he2022masked,xie2022simmim,chen2023context} divide an image into visible and masked patches, and train the model to predict the masked patches from the visible ones.
The prediction targets for visible patches include pixel values~\cite{he2022masked}, HOG features~\cite{wei2022masked}, and high-level semantic features~\cite{wei2022mvp}.
While the main purpose of MIM is representation learning, it has also shown potential for image generation.
For example, Cao \emph{et al.}~\cite{cao2022learning} use MAE features and attention scores to help a convolutional inpainting model capture long-range dependencies.
In contrast, this paper employs MAE priors to enhance the context stability of latent inpainting models.

\paragraph{Image harmonization} aims to blend a foreground object into a background image to get realistic and visually consistent composition~\cite{tsai2017deep}.
This task is often framed as an image translation problem~\cite{zhu2015learning, cong2020dovenet, guo2021intrinsic, cong2022high, niu2023deep, wang2023semi,wang2026next,meng2024high, ren2024relightful}.
In a similar way, our work tackles color inconsistency issues in latent generative models.
The key difference is that, in image harmonization, inconsistencies come from combining images from different sources and therefore different real image distributions, while in latent generative models, color inconsistencies arise from distribution shift of the learned VAE and the generative model.

\paragraph{Object insertion and removal}
are two opposite tasks in image inpainting. 
Object insertion adds new objects to an image via shape-guided masks~\cite{zeng2022shape,xie2023smartbrush}, text prompts~\cite{wang2023imagen,xie2023smartbrush,canberk2024erasedraw,chiu2024brush2prompt}, learnable prompts~\cite{wang2024repositioning,zhuang2024task,chiu2024brush2prompt}, additional network modules~\cite{chen2024improving,ju2024brushnet}, or reference images of objects~\cite{saini2024invi,gong2026direct}.
Some studies also focus on completing partial objects using reference images~\cite{cao2024leftrefill} or learnable prompts~\cite{wang2024repositioning}.
Object removal, in contrast, aims to erase unwanted objects from an image.
Typical methods include attention reweighting~\cite{li2024magiceraser} and learnable prompts~\cite{wang2024repositioning,zhuang2024task}.
These techniques can support dataset creation~\cite{de2024placing}, while new datasets themselves can also help improve these tasks~\cite{winter2024objectdrop}.
Most prior work has focused on building stronger inpainting models.
In contrast, our work addresses a fundamental issue in latent generative models: they often insert unintended objects into the inpainting region.
We also propose solutions to mitigate this problem.

\section{Problem Definition}
\paragraph{Problem setup}
\label{sec:problem-setup}
Inpainting takes a masked image and a mask that marks the missing area to be filled.
The goal is to reconstruct the missing part using the visible regions, producing realistic and high-quality images.
Let $\bm{x}\in[0,1]^{H\times W\times 3}$ be an image, $M\in\{0,1\}^{H\times W}$ a binary mask, and $\bm{x}_{\bar M}=(1-M)\odot\bm{x}$ the visible observation.
An inpainting model produces $\hat{\bm{x}}=\bm{x}_{\bar M}+M\odot G_{\theta}(\bm{x}_{\bar M},M,c,\epsilon)$, where $c$ is optional guidance and $\epsilon$ is the random noise.
We focus on the standard inpainting task \emph{without adding extra text conditions}. 

We address two common problems in inpainting models:

(1) \emph{Unwanted object insertion}: unstable hallucinations that generate random, irrelevant elements in the masked area.
A generated semantic instance $a$ inside $M$ is an \emph{unwanted object} if it is not supported by the visible context or condition, i.e., if $a\notin\mathcal{S}_{\tau}(\bm{x}_{\bar M},M,c)$, where $\mathcal{S}_{\tau}$ denotes supported semantic instances.
The hallucination rate is the expectation rate of such unsupported instances over evaluation dataset.
In practice, this semantic support is estimated by human or VLM judgment.

We attribute unwanted object insertion to three interacting factors.
First, random masking with global captions can train the model to reconstruct fully masked objects even when the caption does not specify them, so some noise patterns become spuriously semantic.
Second, residual cues in the unmasked region, such as shadows, mask shape, or repeated nearby instances, may encourage the model to recreate the removed object.
Third, attention-based generators may not sufficiently prioritize user or context guidance over their learned image prior.
We address these factors post hoc by introducing a context-stable prior derived only from the visible region.

\begin{figure}[t]
\centering
\includegraphics[width=0.9\linewidth]{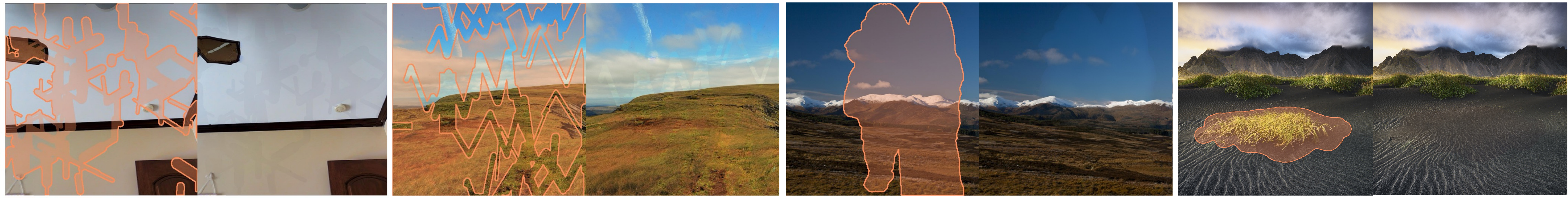}
\caption{
Color shift exists in all kinds of cases, including indoor and outdoor scenes, random or continuous masks, causing darker or lighter color shift.
}
\label{fig:color-shift}
\end{figure}

\begin{figure}[t]
\centering
\includegraphics[width=0.9\linewidth]{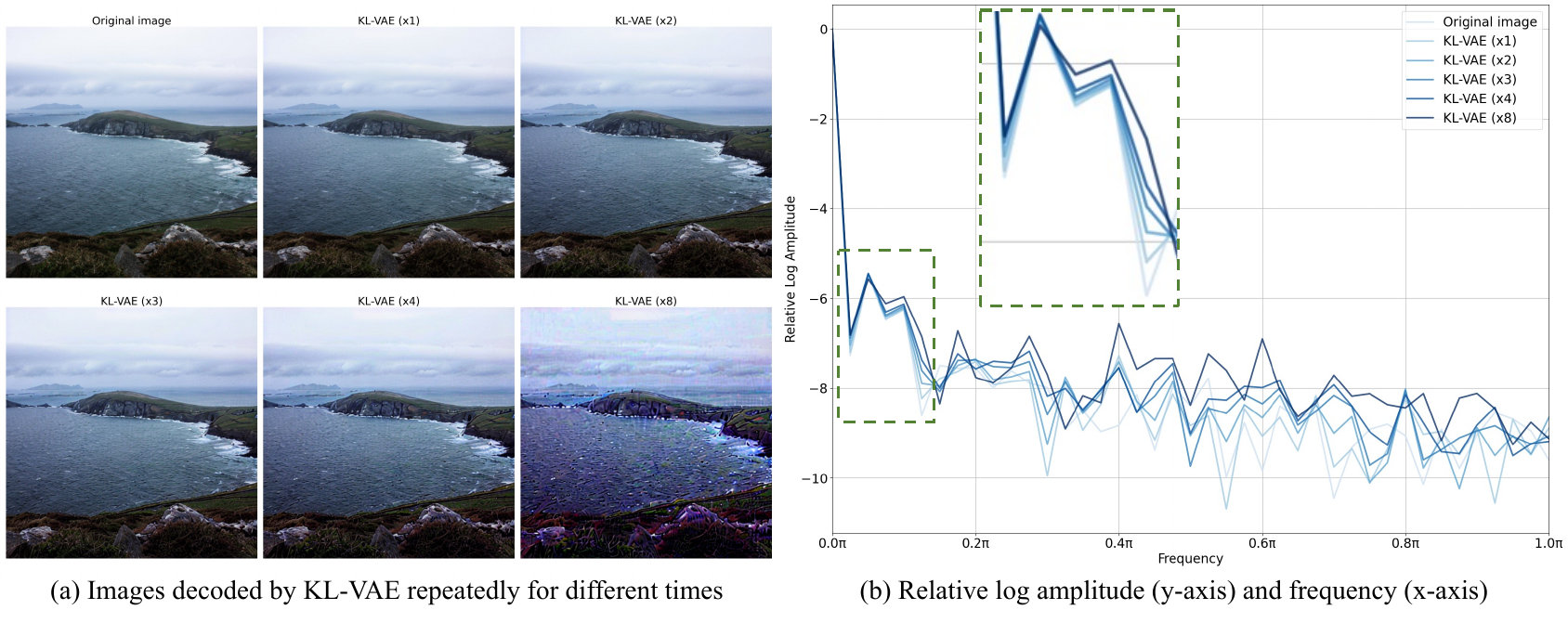}
\caption{
(a) Larger color
shift is observed during repeated reconstruction. (b) VAE suffers from non-ignorable shifts in low-frequency fields.
}
\label{fig:info_loss}
\end{figure}

\begin{figure}[t]
\centering
\includegraphics[width=0.5\linewidth]{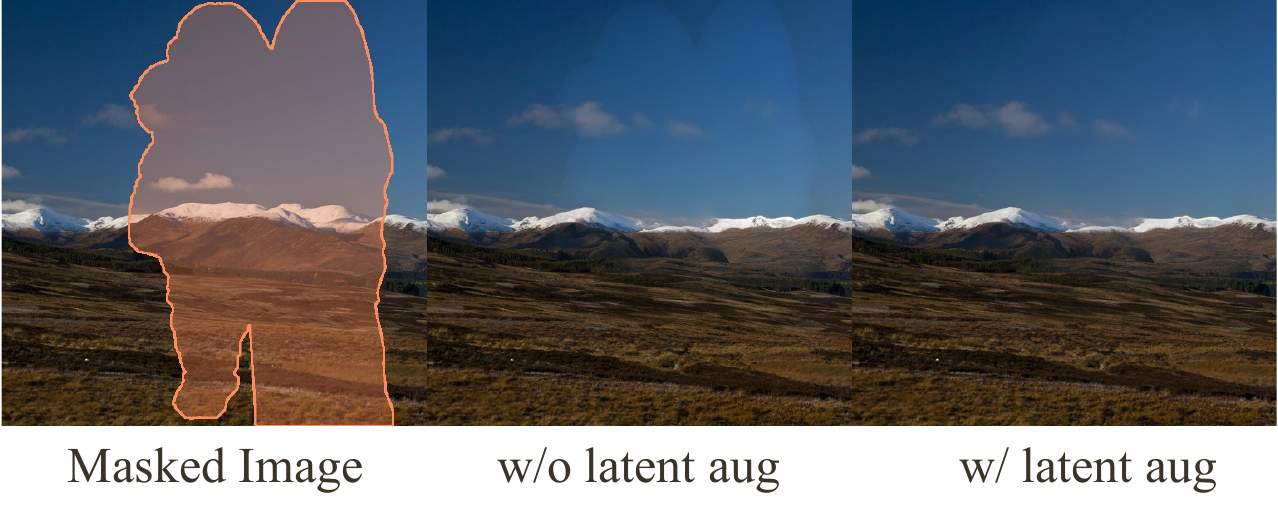}
\caption{
ASUKA w/o latent aug only captures degradation of VAE, thus still has color shift in some cases.
The latent augmentation handles the gap between generated and real latent, further improving the color consistency.
\label{fig:latent-aug-visualization}
}
\end{figure}

(2) \emph{Color inconsistency}: mismatched colors between masked and unmasked regions, often leaving smear-like artifacts.
Mathematically, we define this as low-frequency discontinuity between generated and visible regions near the mask boundary.
Let $L_{\sigma}$ be a low-pass operator and $B_r(M)$ an $r$-pixel band around the mask boundary.
A boundary inconsistency score is
\begin{equation}
\small
  \mathrm{CI}_{r}(\hat{\bm{x}};\bm{x},M)=
\sum_{i\in B_r(M)}
\|\nabla L_{\sigma}(\hat{\bm{x}})_i-\nabla L_{\sigma}(\bm{x})_i\|_1/|B_r(M)|.
\end{equation}
Latent inpainting models first decode the generated latents and then combine them with the ground-truth visible pixels.
However, existing models often produce color inconsistencies that appear as brightness, saturation, hue, or luminance seams along the mask boundaries, as shown in Fig.~\ref{fig:color-shift}.
This problem has two main causes.
(1) The \emph{low-frequency reconstruction loss of the VAE cannot be ignored}, as shown in Fig.~\ref{fig:info_loss}(b). The VAE not only loses high-frequency details but also introduces noticeable color shifts. This becomes more obvious when the VAE is applied repeatedly, as shown in Fig.~\ref{fig:info_loss}(a), where the color shift becomes stronger after each reconstruction.
(2) The gap between the generated latents and the real latents also causes color inconsistency, even when the VAE reconstruction loss is reduced, as shown in Fig.~\ref{fig:latent-aug-visualization}.
To achieve better color consistency, both the VAE reconstruction loss and the latent generator need to be improved.

We provide more detailed definitions in 
Sec.~\ref{sec:supp-problem-definition} 
and analysis of the two issues in Supplementary 
Sec.~\ref{sec:supp-empirical-discuss}.

\paragraph{Backbone Models}
We study two representative latent inpainting models that use a VAE~\cite{kingma2013auto} to compress images into a lower-dimensional latent space:
(1) the U-Net–based Stable Diffusion v1.5 inpainting model (SD)~~\cite{Rombach_2022_CVPR},
(2) the MMDiT-based FLUX.1-Fill-dev inpainting model~\cite{flux-fill}.
In SD, a diffusion process~\cite{sohl2015deep} maps the latent   to Gaussian noise, and a U-Net~\cite{ronneberger2015u} learns the reverse denoising process. 
Text conditioning is introduced through cross-attention layers~\cite{Vaswani2017Attention}. 
For inpainting, SD extends the U-Net input by concatenating the masked image and mask with the noise along the channel dimension.
In contrast, FLUX uses rectified flow~\cite{liu2022flow,albergo2022building,lipman2022flow} to map the latent  to noise and employs a vision transformer~\cite{peebles2023scalable} for generation. 
Text conditioning is applied by concatenating text embeddings with image patches as transformer input, and a pooled text embedding is injected into the normalization layers. 
For inpainting, FLUX.1-Fill-dev~\cite{flux-fill}  also concatenate the masked image and mask in the channel dimension. 

\section{ASUKA}

\begin{figure}
\centering
\includegraphics[width=0.9\linewidth]{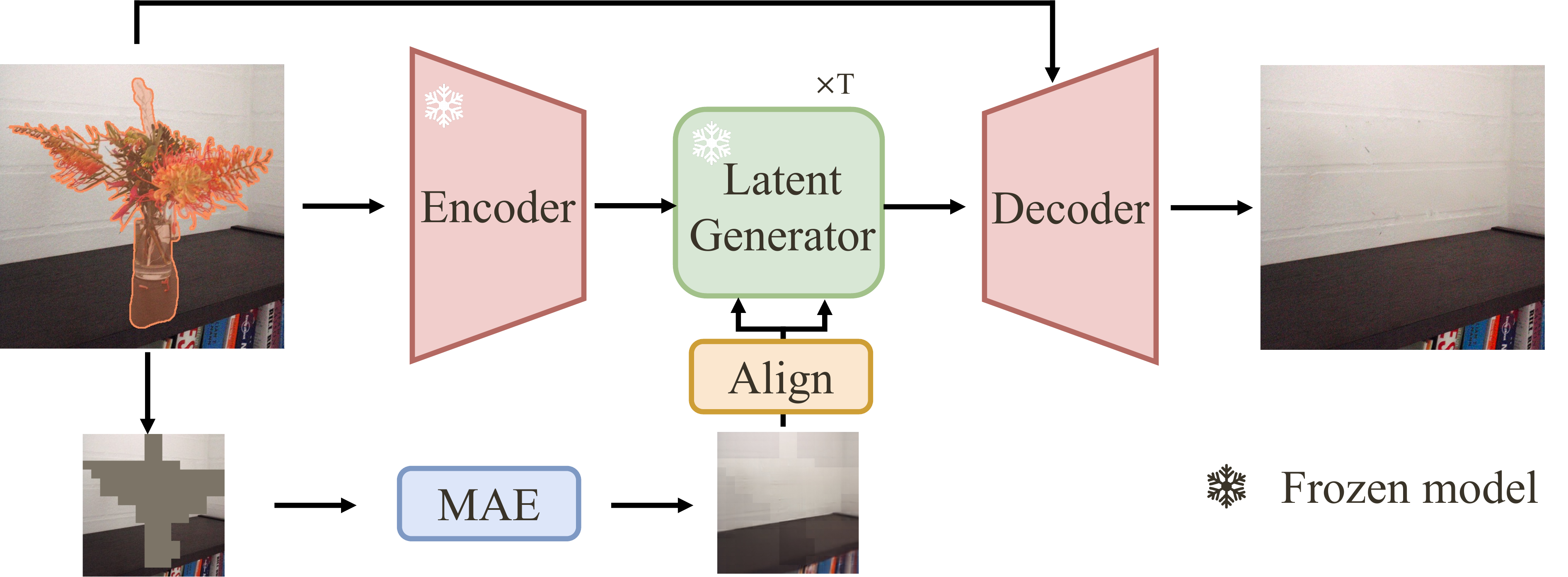}
\caption{
ASUKA tackles the unwanted object insertion issue by adopting the MAE to provide a stable prior for latent generative models to maintain the generation capacity while mitigating object hallucination.
For the color-inconsistency issue, ASUKA utilizes an inpainting-specialized decoder to achieve mask-unmask color consistency when decoding latent.
}
\label{fig:asuka}
\end{figure}

\paragraph{Overview}
The framework of the proposed Aligned Stable inpainting with Unknown Areas prior (ASUKA) is shown in Fig.~\ref{fig:asuka}.
ASUKA builds on pre-trained latent inpainting models.
Our goal is to reduce object hallucination and improve color consistency in inpainting, while preserving the generation ability of frozen models.
ASUKA consists of:
(1) a \emph{context-stable alignment} that aligns the stable Masked Auto-Encoder (MAE) prior for masked regions with generative models, and
(2) a \emph{color-consistent alignment} that aligns the ground-truth unmasked region with the generated masked region during decoding.
To achieve this, we inject our proposed MAE prior as a condition for the generation process to reduce hallucination.
We then introduce an alignment module that connects the MAE prior with the generative models, trained using the same objective as the generative models.
In addition, to handle color inconsistencies caused by the VAE decoder and generative model which often lead to mismatches between masked and unmasked regions, we train a decoder specialized for inpainting. 
This decoder perform the local harmonization task during mapping the latent features back into image space, ensuring seamless color consistency.

In this section, we first present the principled design of our framework.
We then describe the implementation of ASUKA-I, discuss its limitations when applied to MMDiT-based models, and introduce the improvements in ASUKA-II.

\subsection{Mitigate Object Hallucination with Stable Prior}
\paragraph{Context-stable prior}
Recent generative models often rely on random noise to increase diversity in their outputs, but this can also cause unintended objects to appear.
Some inpainting models address this by using reconstruction loss to recover masked regions, but they also add other losses such as perceptual loss~\cite{suvorov2022resolution}, which can reduce stability.
In contrast, MAE is known to provide stable predictions of masked regions by relying only on the visible context.
We use MAE to generate a stable prior, ensuring that \emph{the improvement can be directly linked to better mitigation of object hallucination}.

\paragraph{MAE as a context-stable prior} 
Since MAE is trained with an L2 reconstruction loss, its output can be seen as a mean estimate. This makes it useful for providing a context-stable prior that helps generative models avoid inventing new concepts.
However, mean estimator tends to produce blurry, averaged results and fails to reconstruct fine details in masked regions. It also performs poorly when used directly as the initialization for inpainting models in an image-to-image setup (see 
Fig.~\ref{fig:supp-mae-to-image} 
for an illustration).
Hence, we use MAE as a prior to stabilize diffusion models instead.

\paragraph{Train MAE on realistic inpainting masks}
\label{sec:mask-strategy}
The original MAE is trained on random masks that are uniformly distributed across an image, but inpainting tasks usually involve large, continuous missing regions. 
Inspired by our previous work~\cite{cao2022learning}, we fine-tune MAE for inpainting-specific masks.
To make MAE more suitable for real-world inpainting, we design a systematic masking strategy with three types of mask bases: object-shape, irregular, and regular. 
Object-shape masks are taken from COCO~\cite{lin2014microsoft} object segments. 
Irregular masks are borrowed from prior work, including the Co-Mod mask~\cite{zhao2021large} and LaMa mask~\cite{suvorov2022resolution}. 
Regular masks include rectangles and their complementary regions.
To ensure diversity and generalization, we sample masks with probabilities of 50\% for object-shape, 40\% for irregular, and 10\% for regular. 
Additionally, for object-shape masks, we combine them with irregular masks half of the time. 
This setup reflects the types of masks commonly found in inpainting tasks, such as object removal or user-specified irregular masks.
We control the mask ratio within $[0.1, 0.75]$ to match the MAE training setup. For masks smaller than 75\%, we enlarge them to reach 75\% by randomly expanding the masked regions. This allows ASUKA to better handle challenging large-hole inpainting cases.

\paragraph{Align MAE prior with frozen generator}
Generative inpainting models are usually trained with text conditions, not MAE priors.
Since we do not assume a text condition for the inpainting task, we propose replacing the text condition with our MAE prior.
However, because the generative models are not fine-tuned, they do not naturally align with the MAE prior.
To address this, we introduce an alignment module that bridges MAE and the generative models in both dimensions and distributions, as shown in 
Fig.~\ref{fig:supp-alignment}.

\paragraph{Dimension alignment}
The MAE prior $F_{\mathrm{MAE}}$ has size $N_{m}\times M_{m}$, where $N_{m}$ is the sequence length and $M_{m}$ is the feature dimension.
To align it with the diffusion or flow condition of size $N_{s}\times M_{s}$, we use a linear layer to map the feature dimension from $M_{m}$ to $M_{s}$.
We also set $N_s = N_m$ so that the local structure of the MAE prior is preserved.

\paragraph{Distribution alignment}
Once the dimensions are aligned, we apply self-attention blocks to help the model learn how to use the prior more effectively. This produces the condition $C_{\mathrm{MAE}}$.
We train the alignment module with the standard generation objective, using the same masking strategy as in MAE training, while keeping all other modules fixed.

\paragraph{Handle misalignment}
When training the alignment module with the set (input image, MAE prior, inpaint result), misalignment can occur.
For instance, if an object is completely masked, the MAE predicts the missing area as background, while generative models try to restore the actual object.
This mismatch may cause the alignment module to ignore the MAE prior.
To fix this, we encourage the generative models to follow the MAE prior more closely by replacing the MAE predicted prior with the MAE reconstructed prior with probability $p$.
The MAE reconstructed prior is obtained by running MAE on the full image without masking, so it has access to all information for reconstruction.
This method trains the alignment module to make better use of the guidance.

\subsection{Enhance Color-Consistency in Decoding}

\begin{figure}
\centering
\includegraphics[width=0.7\linewidth]{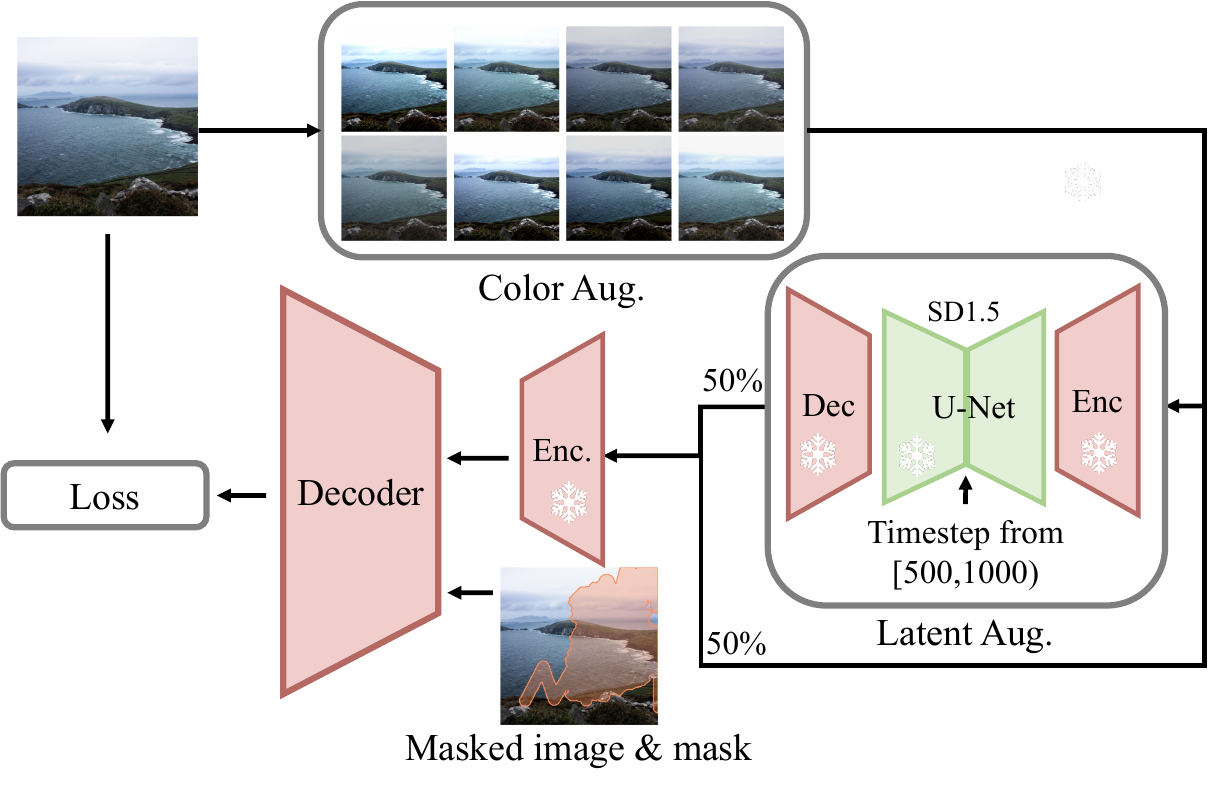}
\caption{
Decoder trained by reconstructing original image from noisy latent augmented in color and latent spaces conditioned on clean unmasked region.\label{fig:aug}}
\end{figure}

\begin{figure}
\centering
\includegraphics[width=0.7\linewidth]{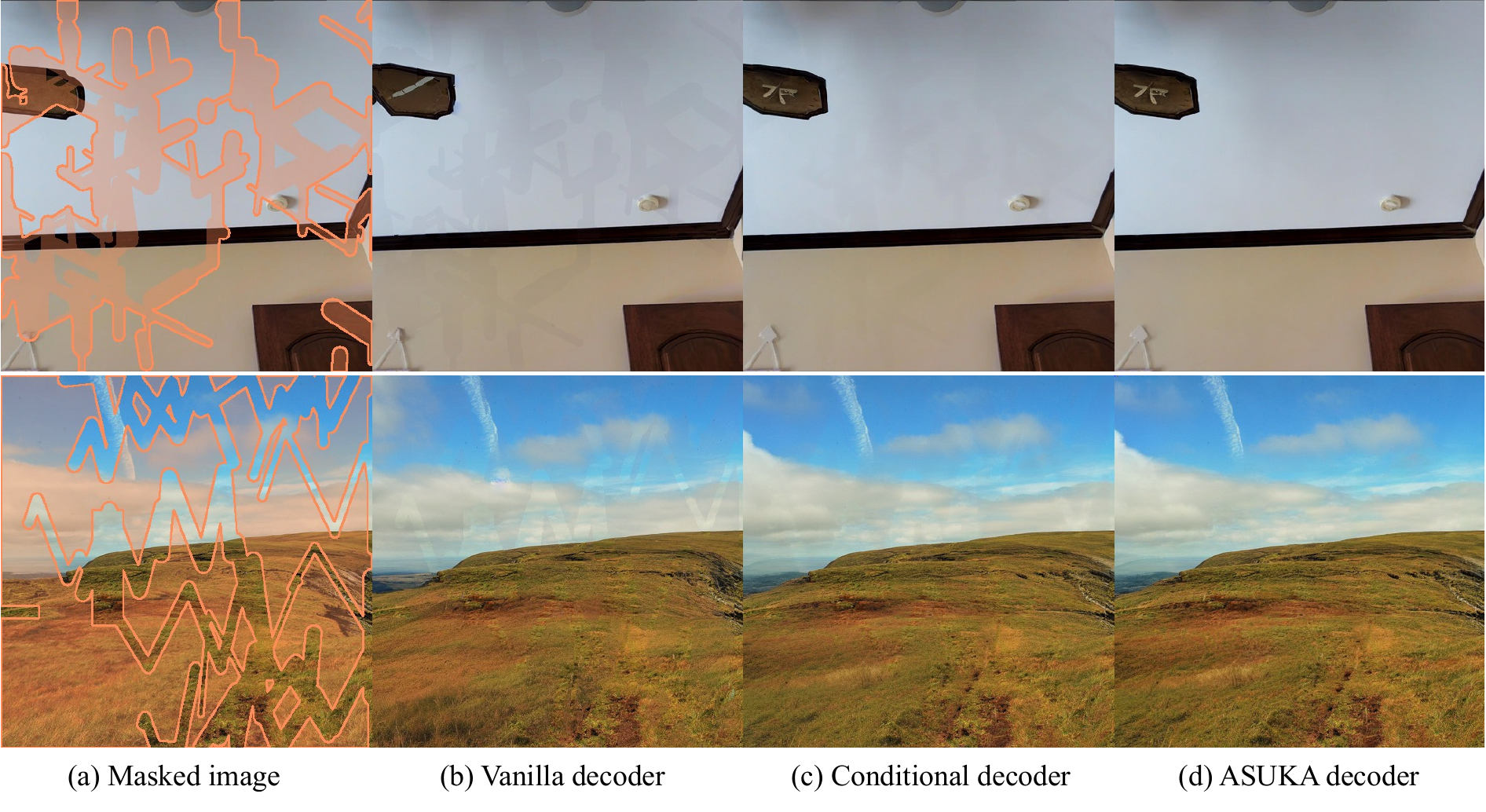}
\caption{SD1.5 inpainting results decoded by (b) vanilla decoder of SD~\cite{Rombach_2022_CVPR}, (c) conditional decoder~\cite{zhu2023designing}, (d) our decoder.
\label{fig:decoder_compare}}
\end{figure}

We propose to solve the color-inconsistency issue and ensure the mask-unmask alignment during VAE decoding.

\paragraph{Unmask-region conditioned decoder}
A straightforward approach is to use the ground-truth unmasked region during decoding, so the model can rely on accurate color information. 
Zhu \emph{et. al.}~\cite{zhu2023designing} follow this idea by adding the masked image as an extra input to the decoder.
However, this method still struggles when the colors and textures of the original and reconstructed images do not match, especially in complex scenes (see Fig.~\ref{fig:decoder_compare} (c)). 
The gap between degraded and original images makes it difficult to directly solve this problem.

\paragraph{Mask-unmask color-consistent decoder}
To make the decoder preserve color consistency between the generated latent and unmasked pixels, we reformulate decoding as a local harmonization task. 
Our decoder takes both the masked image (in pixel color space) and the binary mask as inputs.

For training, we design a color and latent augmentation strategy (see Fig.~\ref{fig:aug}) to expose and amplify color inconsistencies. 
We follow the standard VAE training setup, but replace the inputs with augmented versions. 
Specifically, the original image is used as the reconstruction target, while the input image is corrupted through:
(1) Color augmentation: simulating color distortions from the VAE.
(2) Latent augmentation: simulating domain gaps between generated and real latents.
This setup forces the decoder to reconstruct a clean, consistent image while aligning with the ground-truth unmasked regions.

\paragraph{Color augmentation} 
We apply color augmentation to measure the VAE loss.
In practice, conditioning on the unmasked image helps reduce but does not fully fix the color inconsistency problem (see Fig.~\ref{fig:decoder_compare} (c)).
Therefore, we explicitly train the decoder to maintain color consistency.
Specifically, we augment each training image by changing its brightness, contrast, saturation, and hue, and then require the decoder to reconstruct the original image using the unmasked image as a reference.
This encourages the decoder to accurately follow the colors in unmasked regions.

\paragraph{Latent augmentation} 
To simulate the gap between generated and real latents, we add artifacts from generative models into the decoder’s training.
However, repeatedly denoising back to real images is very time-consuming, even when using DDIM~\cite{song2021denoising}.
To balance speed and accuracy, we design a one-step estimation method.
Since our goal is to capture the generation gap, we condition on the clean latent $\bm{z}_0$ and an all-zero mask $\mathbf{O}$.
This gives the generator all the information needed to produce the clean latent, ensuring the source of degradation come from the learned latent shift.
Following standard pipeline, we estimate $\bm{z}_0$ under these modified conditions:
\begin{equation}
\hat{\bm{z}}_0=\frac{1}{a}(\bm{z}_t-b\varepsilon_\theta\left([\bm{z}_t;\bm{z}_0;\mathbf{O}],t\right)),
\label{eq:z0_pred}
\end{equation}
where the timestep $t$ is randomly sampled from $[500,1000)$.
Here, $a$ follows the prescribed variance schedule, with $a^2+b^2=1$ in diffusion models and $a+b=1$ in rectified flow models.
$\varepsilon_\theta(\cdot)$ is the frozen generator, which takes as input the noised $\bm{z}_t$, the unmasked $\bm{z}_0$, and the all-zero mask $\mathbf{O}$.

We use large-step denoising to enlarge the distribution gap, since empirically the generator produces stable results at small $t$ when conditioned on the unmasked latent $\bm{z}_0$.
The estimated latent $\hat{\bm{z}}_0$ is then decoded into an image, serving as augmented input.
This makes latent augmentation an offline strategy.
We apply it to 50\% of the training images.
The fine-tuned decoder achieves noticeably better consistency, as shown in Fig.~\ref{fig:decoder_compare}.

\subsection{ASUKA-I}

In our conference version~\cite{wang2025towards}, we instantiate the ASUKA framework on the SD~1.5 inpainting model~\cite{Rombach_2022_CVPR}, which adopts a U-Net–based latent diffusion backbone.  
This section briefly reviews the concrete design choices of ASUKA-I and clarifies its scope and limitations, which directly motivate the improved ASUKA-II framework introduced later.

\paragraph{Replace text condition with MAE prior}
In SD 1.5, text features are injected into each cross-attention layer to guide the generation process, using a fixed and shared text representation.
ASUKA-I replaces the text condition with the proposed MAE prior, which is directly fed into the same cross-attention layers.
This design ensures minimal modification to the original architecture while enabling context-stable structure-aware control through the MAE representation.

\paragraph{Efficient low-resolution decoder fine-tuning}
To enhance color consistency, ASUKA-I fine-tunes the SD VAE decoder using the augmented training strategy at a low resolution of $256^2$.
Despite being trained at low resolution, the decoder generalizes well to $512^2$.
This strategy significantly reduces training cost while maintaining strong decoding quality and improved color consistency.

\paragraph{Extending ASUKA-I to transformer-based generators.}
With the rapid adoption of transformer-based image generation models, in this manuscript, we further apply ASUKA-I to the multimodal transformer-based inpainting model, FLUX-1-Fill-dev~\cite{flux-fill}.
While ASUKA-I-FLUX consistently outperforms the original FLUX-1-Fill-dev, this extension exposes several limitations of ASUKA-I:
(1) \emph{Complex text conditioning}:  
Unlike U-Net models, MMDiT-style architectures treat text as part of the model input rather than a fixed condition injected through cross-attention.  
The deep transformation of text features makes them difficult to align with MAE representations, resulting in unstable conditioning.
(2) \emph{Position modeling.}  
Transformer-based models require explicit positional information for conditioning inputs.  
Without proper position modeling, controllability is significantly reduced.
(3) \emph{Limited decoder generalization.}  
The fine-tuned decoder in ASUKA-I struggles to generalize to realistic, high-resolution masks commonly encountered in real-world applications.

These observations highlight that ASUKA-I, while effective for U-Net–based diffusion models, does not fully generalize to modern transformer-based generators, motivating the systematic redesign for ASUKA framework.

\subsection{ASUKA-II}
Our ASUKA-II systematically extends ASUKA-I with context-stable transformer alignment and refined color-consistent training, yielding significantly more robust suppression of unwanted object insertions and improved visual coherence. In this section, we elaborate these improvement.

\paragraph{Overview}
\label{sec:mae-inte}
(1) To achieve more context-stable alignment, we integrate MAE features into every transformer layer using LoRA and gating mechanisms for finer control, avoiding the need to fit the complex text-conditioning component in MMDiT.
We further strengthen spatial correspondence through image-based positional encoding and resolution-aware scaling.
(2) To enhance color-consistent alignment, we refine the training recipe to reduce boundary artifacts, a common issue in real-world applications.
Combined together, these enhancements significantly improve the inpainting performance of ASUKA-II-FLUX over ASUKA-I-FLUX, producing more context-stable and visually coherent results.

\paragraph{Over-complex text condition transformation}
In ASUKA-I, we replace the text condition with our proposed MAE condition. While this method works effectively under the cross-attention framework in U-Net based SD 1.5, we found that directly substituting the text input in FLUX with the MAE condition fails to control the generation process properly.
This different behavior is caused by how the two architectures handle conditioning.
In cross-attention–based models (SD 1.5), the text condition remains fixed and shared across all layers, allowing each layer to extract relevant information independently. 
In contrast, in MMDiT used by FLUX, the text condition is treated as input, continuously transformed to align better with the evolving image representation. While this improves internal coherence, it also reduces the model’s ability to generalize to alternative conditioning signals such as our MAE prior.
Therefore, directly injecting the MAE prior as an input to FLUX is not ideal.
Inspired by the success of per-layer MAE feature injection in SD 1.5, we propose a similar per-layer integration strategy for FLUX.

\begin{figure}
\centering
\includegraphics[width=0.7\linewidth]{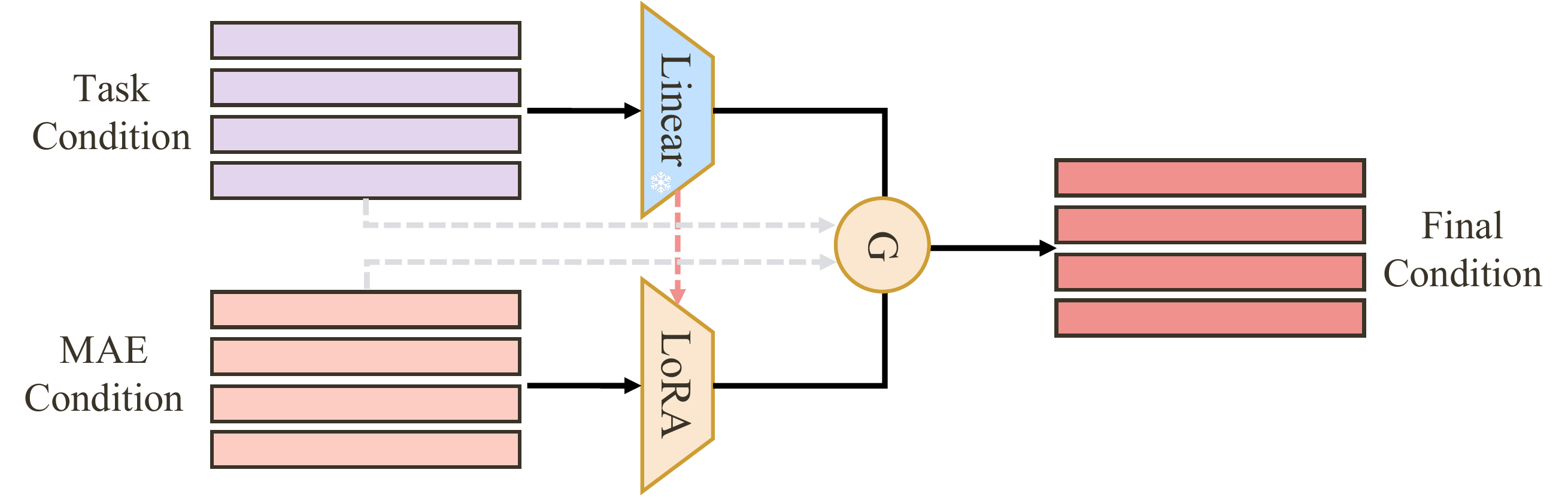}
\caption{
Conditioning mechanism in ASUKA-II-FLUX.
We replace the text condition with a learnable task condition.
For the shared MAE condition, we use a LoRA linear layer with gated fusion module to generate the condition QKV.
This design adds no extra attention computation.
\label{fig:mae-integration}}
\end{figure}
\paragraph{Per-layer MAE condition integration} 
To maintain the original capability of the FLUX model, we do not completely remove the text condition.
Instead, following our SEELE work~\cite{wang2024repositioning}, we introduce learnable inpainting-specific task prompts to replace the original text input.
We process the MAE feature using a shared alignment module, as in ASUKA-I, to obtain the MAE condition $f_{mae}$. 
For each layer, we then integrate this MAE condition with the task condition to achieve per-layer MAE control.
Concretely, we introduce a LoRA module into the QKV layers of the task condition, denoted as $\Delta W_{q,k,v}$, to learn MAE-adapted representations. 
We sum the MAE and task QKV outputs to form the final condition, with their relative contributions modulated by a gating module $G_{q,k,v}$. 
The gate consists of a linear layer followed by a sigmoid activation, which adaptively balances the two sources.
For each layer $i$ and component $j \in \{q,k,v\}$, we have:
\begin{equation}
\begin{aligned}
f_{mae, j}^{i} &= (W_{j}+\Delta W_{j})f_{mae}^{i},\quad f_{task, j}^{i} = W_{j}f_{task}^{i}, \\
f_{j}^{i} &= f_{task, j}^{i} + G_{j}(f_{mae} + f_{task}^{i}) \odot f_{mae, j}^{i}.
\end{aligned}
\end{equation}
An illustration of the integration is shown in Fig.~\ref{fig:mae-integration}.
This design modifies the conditional features only through lightweight LoRA adapters, keeping the added computational cost minimal.
Moreover, since we do not increase the sequence length, the attention computation complexity remains unchanged.
Note that since we only change the conditioning features, the overall backbone still remain freeze.

\paragraph{Spatially-Aware Positional Encoding}
In transformer models, positional encoding helps the network understand the spatial relationships between tokens. 
Although MAE already applies positional encoding, this information is often not effectively utilized by FLUX, especially at higher resolutions.
To better align the spatial correspondence between MAE features and the noise latent, inspired by~\cite{zhang2025easycontrol}, we apply a scaled positional encoding to the final condition.
We initialize the positional IDs using the original image position IDs from FLUX. Because MAE features are extracted from lower-resolution images, we scale the spatial dimensions of $image\_ids$ by a factor $S = \frac{R_{img}}{R_{mae}}$, where $R_{img}$ is the latent image resolution and $R_{mae}$ is the MAE condition resolution.
The final positional encodings of the condition are given by $(0,0), (0,S), \ldots, (P \times S, P \times S)$ where $P$ denotes the number of MAE patches.
This scaled positional encoding ensures a more accurate and fine-grained spatial alignment between the MAE features and the noise latent representation.

\begin{figure}
\centering
\includegraphics[width=0.7\linewidth]{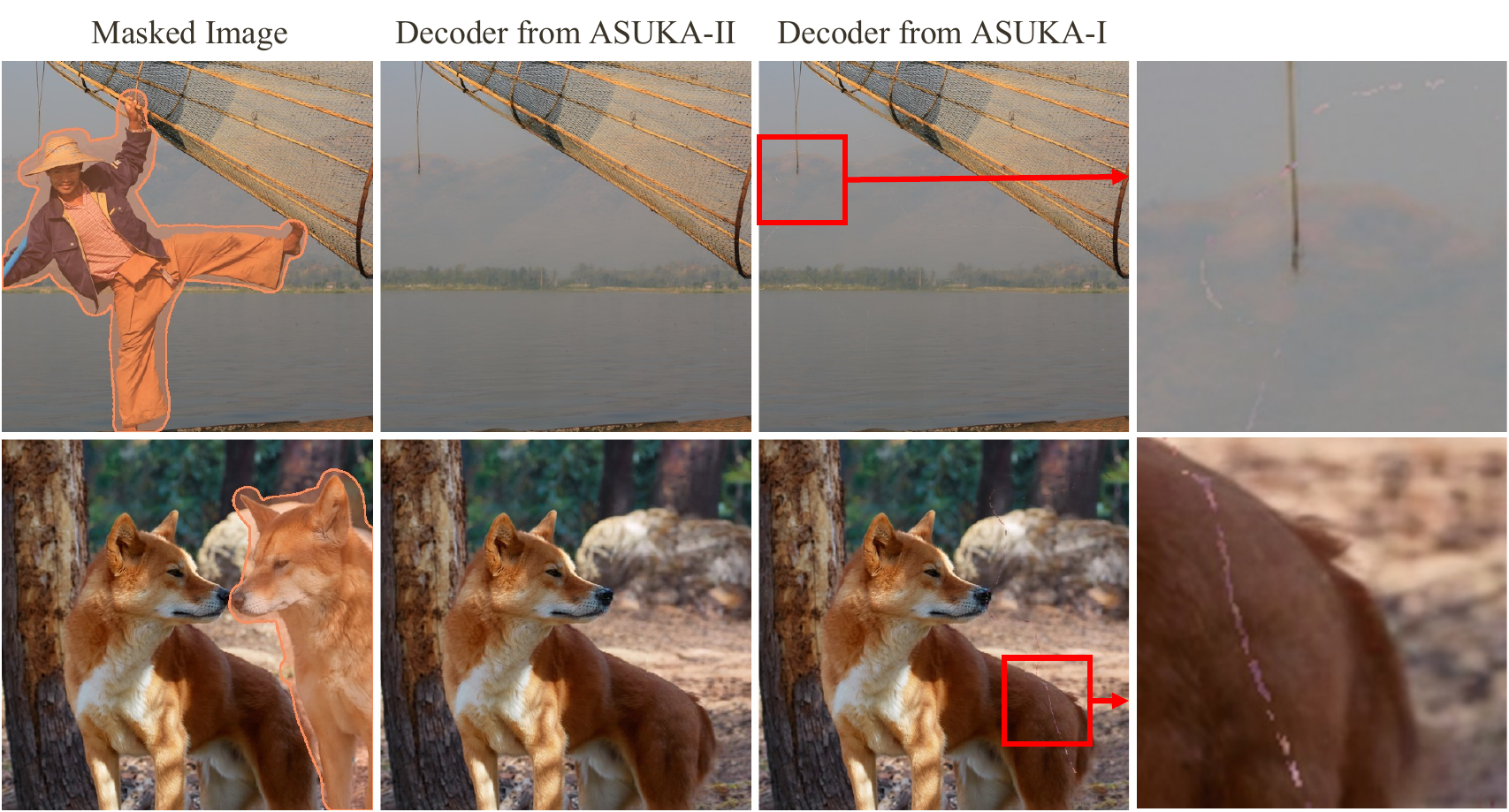}
\caption{
Jagged masks, caused by downsampling from high-resolution masks, often lead to visible boundary artifacts along mask edges.
ASUKA-II solves this issue and produces more realistic inpainting results.
\label{fig:jagged-mask}}
\end{figure}

\paragraph{Improved Mask shape coverage: Simulated Jagged Masks}
\label{sec:aug-decoder}
In ASUKA-I, we simulate masks using a combination of object-shaped masks, irregular masks, and regular masks. 
These mask types are designed to cover a wide range of plausible occlusion patterns and work well in most settings. 

However, in practice, we observed that an important class of masks is not adequately represented.
In real-world inpainting applications, users often start from high-resolution images and carefully annotated masks with precise boundaries. 
Due to hardware or memory constraints, both the image and the mask are usually resized to a lower resolution before being fed into the inpainting model. 
When nearest-neighbor or similar discrete resizing operations are applied, this process introduces jagged and staircase-like boundaries in the mask. 

In our preliminary experiments, we found that these jagged masks primarily affect the decoding stage, leading to visible boundary artifacts and degraded visual consistency near mask edges, as illustrated in Fig.~\ref{fig:jagged-mask}. 
This training-inference mask distributions mismatch limits the robustness of the decoder.

To address this issue, we augment the decoder fine-tuning data by explicitly simulating jagged masks. 
Specifically, we first generate masks at a higher spatial resolution and then down-sample them to the training resolution using nearest-neighbor interpolation. 
This procedure naturally produces jagged mask boundaries that closely resemble those encountered in practical use. 
By exposing the decoder to such masks during training, ASUKA-II learns to better handle boundary discretization effects, resulting in improved decoding quality and reduced artifacts under real-world inpainting scenarios.

\subsection{Theoretical Perspective: Learned Posterior Regularization}
\label{sec:analysis}
We provide theoretical analysis on the problems we focused and why our solutions are theoretically grounded.

\paragraph{MAE as a conditional-mean prior}
Image inpainting is intrinsically ill-posed~\cite{engl2014inverse}: the visible region $\bm{x}_{\bar M}$ and the mask $M$ do not uniquely determine the missing content $\bm{x}_{M}$.
Many completions can be visually plausible, and a generative model therefore samples from a conditional distribution,
\begin{equation}
p_{\theta}(\bm{z}_{M}\mid \bm{z}_{\bar M}, M, c),
\end{equation}
where $\bm{z}=E(\bm{x})$ is the latent and $c$ denotes the conditioning signal.
This formulation is powerful but also risky: if the conditional distribution contains modes that are plausible under the global image prior but weakly supported
by the visible context, sampling may produce unwanted object insertion.
In contrast, an MAE trained with squared reconstruction loss estimates the Bayes predictor,
\begin{equation}
\small
r_{\phi}^{*}(\bm{x}_{\bar M},M)
=\textrm{argmin}_r\mathbb{E}\|\bm{x}_M-r(\bm{x}_{\bar M},M)\|_2^2 
=\mathbb{E}[\bm{x}_M\mid\bm{x}_{\bar M},M].
\label{eq:conditional_mean}
\end{equation}
Thus, the MAE prior is a context-derived conditional mean: it is conservative and stable because unsupported semantic modes are averaged out unless they are strongly implied by the visible region, but it can be blurry because when several completions are possible, the conditional mean averages over them.
This is precisely why ASUKA does not directly use MAE as the final output, but uses it as a stabilizing prior for a high-fidelity latent generator.

This interpretation is also consistent with the theory of denoising autoencoders.
Denoising autoencoders are known to learn directions toward high-density regions of the data distribution, and their reconstruction residual approximates the data
score under small corruption noise~\cite{vincent2011connection,alain2014regularized}.
Although MAE uses masking rather than Gaussian corruption, the same principle applies conceptually: reconstruction from corrupted observations learns a context-conditioned projection back toward the natural image manifold.

\paragraph{Posterior regularization and hallucination suppression}
ASUKA can be interpreted as sampling from a regularized posterior distribution.
Let $\bm{x}_{\phi}=r_{\phi}(\bm{x}_{\bar M},M)$ be the MAE prior.
We define a soft consistency energy,
\begin{equation}
d(\bm{z}_{M},\bm{x}_{\phi})^2
=
\left\|
P_M D(\bm{z}) - \bm{x}_{\phi}
\right\|_2^2,
\end{equation}
where $D$ is the decoder and $P_M$ selects the masked region.
The ASUKA-guided distribution can then be written as:
\begin{equation}
\small
p_{\theta,\lambda}(\bm{z}_{M}\mid \bm{z}_{\bar M},M,\bm{x}_{\phi})
\propto
p_{\theta}(\bm{z}_{M}\mid \bm{z}_{\bar M},M)
\exp
\left(
-\lambda d(\bm{z}_{M},\bm{x}_{\phi})^2
\right).
\label{eq:asuka_regularized_distribution}
\end{equation}
This is a product of the original generative prior and a learned context-consistency likelihood.
It is analogous to posterior sampling in diffusion inverse problems, where the reverse process combines a generative score with a data-consistency or likelihood term~\cite{song2021scorebased,chung2023diffusion,kawar2022denoising,lugmayr2022repaint}.
Here, the MAE prior plays the role of a learned pseudo-measurement of what the visible context supports.
Eq.~\eqref{eq:asuka_regularized_distribution} gives a direct explanation of hallucination suppression.
Consider two sets of completions:
\begin{equation}
\small
\mathcal{A}_{r}=\{\bm{z}_{M}: d(\bm{z}_{M},\bm{x}_{\phi})\le r\},
\quad
\mathcal{H}_{\tau}=\{\bm{z}_{M}: d(\bm{z}_{M},\bm{x}_{\phi})\ge \tau\},
\end{equation}
where $\mathcal{A}_{r}$ contains completions close to the MAE prior and $\mathcal{H}_{\tau}$ contains far-away completions, with $\tau>r$.
Under the regularized distribution,
\begin{equation}
\frac{
p_{\theta,\lambda}(\mathcal{H}_{\tau})
}{
p_{\theta,\lambda}(\mathcal{A}_{r})
}
\le
\exp\left[-\lambda(\tau^2-r^2)\right]
\frac{
p_{\theta}(\mathcal{H}_{\tau})
}{
p_{\theta}(\mathcal{A}_{r})
}.
\label{eq:hallucination_bound}
\end{equation}
Therefore, modes far from the MAE prior are exponentially down-weighted relative to modes close to the context-supported prior.
If unwanted inserted objects correspond to completions that deviate from the conditional-mean background estimate, ASUKA suppresses them by construction.

\paragraph{Diversity-stability trade-off}
ASUKA reduces diversity that is not justified by the visible context, but it does not collapse all stochasticity.
The MAE prior constrains the coarse semantic direction of generation, while texture, fine structure, and local appearance are still sampled by the frozen diffusion or flow model.
Equivalently, ASUKA suppresses context-unsupported semantic diversity, such as arbitrary foreground objects, while preserving context-supported appearance diversity that can be derived from the surrounding scene.

Additional derivations, including the diversity-stability derivative, amortized guidance objective, ASUKA-II injection interpretation, and decoder-as-denoising view, are provided in Supplementary 
Sec.~\ref{sec:supp-theoretical-discuss}.

\begin{figure*}
\centering
\includegraphics[width=0.8\linewidth]{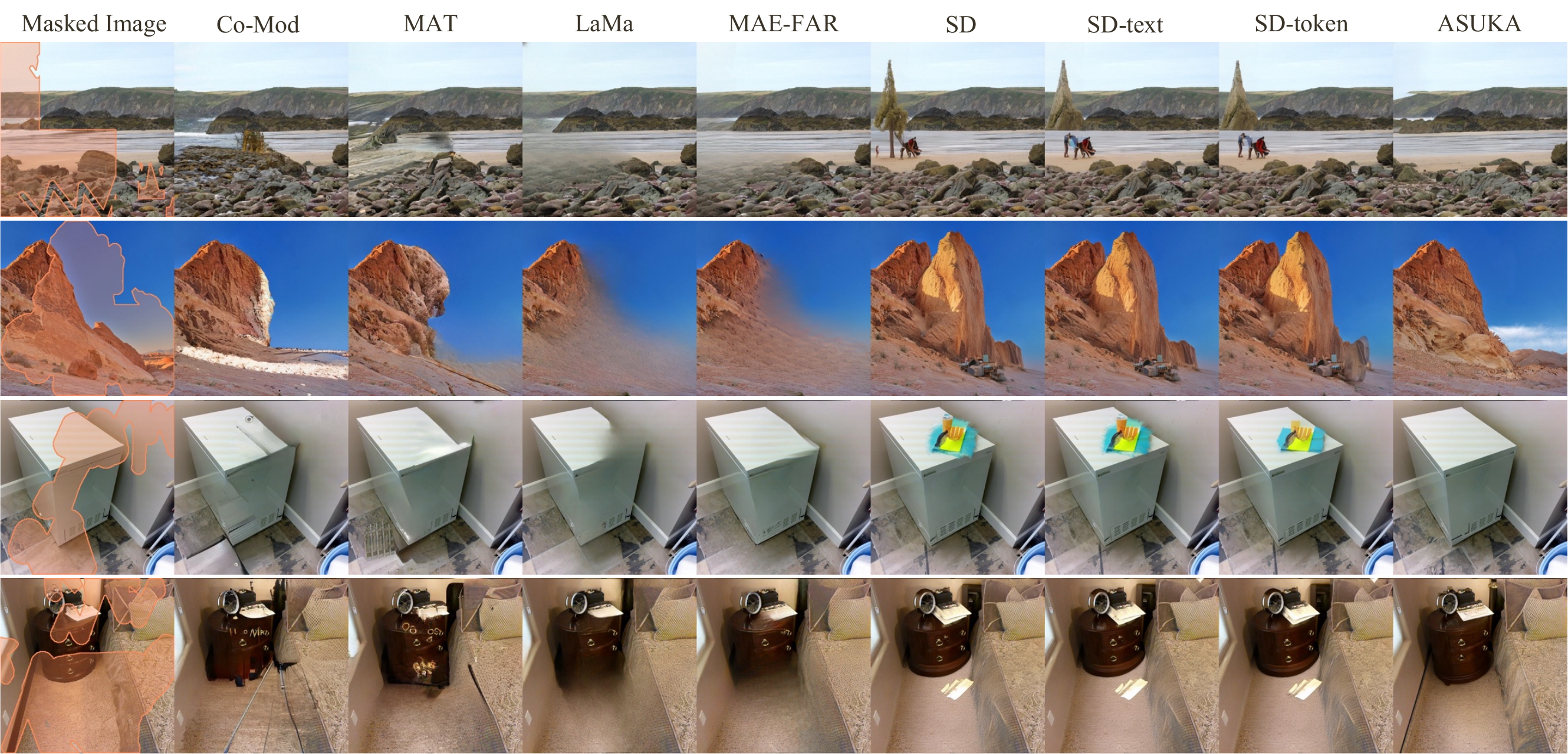}
\caption{
Inpainting results for 512$^2$ images.
GANs generate blurred results; SD variants hallucinate unreasonable objects and suffer from color shift.
ASUKA-I-SD achieves unwanted-object-mitigated and color-consistent inpainting.
}
\label{fig:main-result}
\end{figure*}

\begin{figure*}
\centering
\includegraphics[width=0.8\linewidth]{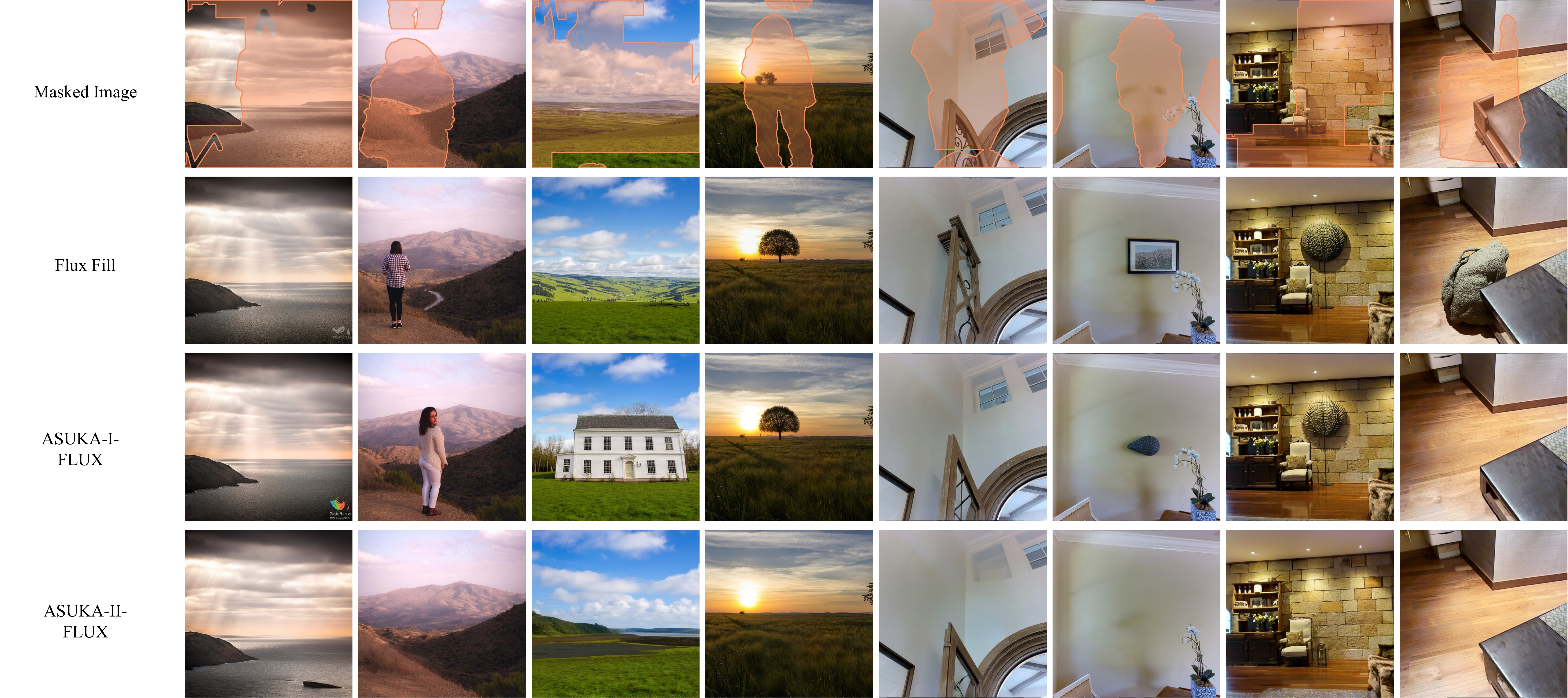}
\caption{
Our ASUKA-II largely improves the unwanted object insertion behavior compared with FLUX-Fill and ASUKA-I.
}
\label{fig:flux-visual}
\end{figure*}

\section{Experiments}

\paragraph{Implementation Details}
We use Places2~\cite{zhou2017places} to train ASUKA. 
For the MAE~\cite{he2022masked} used in ASUKA, 
we train on images of size $256^2$, which is efficient and produce context-stable guidance for generative models to generate high-resolution images.
We fine-tune the MAE with a batch size of 1024.
We train the alignment module with AdamW~\cite{loshchilov2018decoupled} of learning rate 5e-2 with the standard diffusion objective.
We set $p$ as $100\%$ and linearly decay it to $10\%$ in the first 2K training steps and then freeze.
We train the alignment module for 25K steps for both SD and FLUX variants.
For SD's decoder, we fine-tune from~\cite{zhu2023designing} for 50K steps with a batch size of 40 and learning rate of 8e-5 with cosine decay. 
For FLUX's decoder, we fine-tune from the original decoder with the same setup.
We use ColorJitter for color augmentation, with brightness 0.15, contrast 0.2, saturation 0.1, and hue 0.03. 

\paragraph{Evaluation datasets}
\label{sec:dataset}
We follow previous works to evaluate on the standard benchmark Places 2~\cite{zhou2017places} validation set of 36,500 images.
In addition, to validate across different domains and mask styles, we construct a evaluation dataset, dubbed as MISATO, from \underline{M}atterport3D~\cite{Matterport3D}, 
Fl\underline{i}ckr-Land\underline{s}cape~\cite{lin2021infinitygan},
Meg\underline{a}Dep\underline{t}h~\cite{MDLi18}, and C\underline{O}CO 2014~\cite{lin2014microsoft} to handle indoor, outdoor, building and background inpainting, respectively.
We select 500 representative examples of size $512^2$  and $1024^2$ from each dataset, forming a total of 2,000 testing examples.
We adopt the masking strategy as in Sec.~\ref{sec:mask-strategy} with mask ratio of $[0.2, 0.8]$ but excluding the rectangle and complement rectangle masks.
We provide the construction and visualization of the dataset in 
Sec.~\ref{sec:supp-dataset}.

\paragraph{General evaluation metrics}
We use the
Learned Perceptual Image Patch Similarity (LPIPS)~\cite{zhang2018unreasonable} to calculate the patch-level image distances,
Fr\'echet Inception Distance (FID)~\cite{heusel2017gans} to compare the distribution distance, and Paired/Unpaired Inception Discriminative Score (P-IDS/U-IDS)~\cite{zhao2021large} to measure the human-inspired linear separability.

\paragraph{Object hallucination evaluation}
We adopt a VLM-based evaluation protocol to explicitly judge the presence of object hallucination.
We leverage a large-scale vision--language model, \texttt{Qwen3-VL-235B-A22B-Thinking}, to perform pairwise reasoning over inpainting results.
The input to the model consists of a composite image, where the left panel shows the masked input image with a semi-transparent mask (rather than a fully blacked-out region), and the right panel contains the inpainting results produced by different methods.
The VLM is prompted to determine whether hallucinated objects appear in the inpainted region, enabling a more semantically grounded and human-aligned assessment of hallucination behavior.

\paragraph{Color-consistency evaluation}
To assess color consistency, we propose \emph{Gradient@edge} (G@e), which measures the average pixel gradient difference along the boundary of the masked region with respect to the ground-truth image.
This metric captures the smoothness of color transitions between the inpainted area and its surrounding context.
A lower G@e value indicates more consistent color continuity and less noticeable color shift, making it a reliable and complementary measure for evaluating inpainting quality.

\begin{table*}
\centering
\renewcommand\tabcolsep{2.0pt}
\begin{threeparttable}

\caption{Quantitative comparison on MISATO@512 and Places 2.
We compare SD-based methods and FLUX-based models separately.
}

\small{
\begin{tabular*}{\textwidth}{@{\extracolsep{\fill}}lcccccccccc}
\toprule
Dataset & \multicolumn{5}{c}{MISATO@512}  & \multicolumn{5}{c}{Places 2 }  \\
Method 
& LPIPS$\downarrow$ & FID$\downarrow$ & U-IDS$\uparrow$ & P-IDS$\uparrow$ & G@e$\downarrow$
& LPIPS$\downarrow$ & FID$\downarrow$ & U-IDS$\uparrow$ & P-IDS$\uparrow$ & G@e$\downarrow$ \\
\midrule
\midrule

Co-Mod~\cite{zhao2021large} 
& 0.179 & 17.421 & 0.243 & 0.109 & 52.106 
& 0.267 & 5.794 & 0.274 & 0.096 & 166.914 \\

MAT~\cite{li2022mat} 
& 0.176 & 17.261 & 0.255 & 0.122 & 48.722 
& 0.202 & 3.765 & 0.348 & 0.195 & 163.442 \\

LaMa~\cite{suvorov2022resolution} 
& 0.155 & 15.436 & 0.260 & 0.135 & 46.270 
& 0.202 & 6.693 & 0.247 & 0.050 & 153.653 \\

MAE-FAR~\cite{cao2022learning}  
& \textbf{0.142} & 13.283 & 0.282 & 0.153 & \textbf{43.613} 
& \textbf{0.174} & 3.559 & 0.307 & 0.105 & 149.843 \\

SD-Repaint~\cite{lugmayr2022repaint} 
& 0.227 & 27.861 & 0.016 & 0.007 & 80.410 
& 0.251 & 12.466 & 0.217 & 0.045 & 176.421 \\

SD~\cite{Rombach_2022_CVPR} 
& 0.168 & 12.812 & 0.345 & 0.211 & 63.844 
& 0.193 & 1.514 & 0.375 & 0.207 & 160.705 \\

SD-text 
& 0.164 & 12.603 & 0.337 & 0.207 & 63.776 
& 0.191 & 1.506 & 0.373 & 0.202 & 160.418 \\

\midrule

SD-token~\cite{wang2024repositioning} 
& 0.160 & 12.517 & 0.331 & 0.204 & 61.700 
& 0.189 & 1.477 & 0.390 & 0.234 & 158.924 \\

SD-IP~\cite{ye2023ip-adapter} 
& 0.157 & 12.204 & 0.398 & 0.242 & 62.704 
& 0.186 & 1.539 & 0.389 & 0.173 & 148.571 \\

SD-T2I~\cite{mou2024t2i} 
& 0.166 & 13.806 & 0.365 & 0.222 & 63.866 
& 0.195 & 1.720 & 0.384 & 0.160 & 148.549 \\

SD-CAEv2~\cite{zhang2023cae} 
& 0.157 & 29.179 & 0.193 & 0.045 & 69.890 
& 0.192 & 6.887 & 0.287 & 0.065 & 151.863 \\

SD-LaMa~\cite{suvorov2022resolution} 
& 0.157 & 12.159 & 0.390 & 0.256 & 62.726 
& 0.188 & 1.522 & 0.389 & 0.168 & 148.461 \\

\midrule

ASUKA-I-SD 
& 0.150 & \textbf{11.495} & \textbf{0.423} & \textbf{0.312} & 47.753 
& 0.183 & \textbf{1.230} & \textbf{0.413} & \textbf{0.287} & \textbf{147.733} \\

\midrule
\midrule

FLUX-Fill 
& 0.156 & 12.170 & 0.353 & 0.194 & 69.428 
& 0.178 & 1.472 & 0.403 & 0.227 & 69.317 \\

PixelHacker
& 0.151 & 12.768 & 0.342 & 0.165 & 54.420
& 0.179 & 2.350 & 0.386 & 0.198 & \textbf{57.889} \\

OmniPaint
& 0.184 & 13.977 & 0.363 & 0.203 & 79.449
& 0.197 & 1.738 & 0.404 & \textbf{0.252} & 72.867 \\

OmniEraser
& 0.276 & 31.758 & 0.128 & 0.026 & 153.541
& 0.273 & 10.309 & 0.205 & 0.033 & 157.879 \\

\midrule
ASUKA-II-FLUX 
& \textbf{0.139} & \textbf{10.681} & \textbf{0.383} & \textbf{0.241} & \textbf{48.793} 
& \textbf{0.174} & \textbf{1.330} & \textbf{0.410} & 0.241 & 63.517 \\

\bottomrule
\end{tabular*}
}

\label{tab:quantitative}
\end{threeparttable}
\end{table*}

\begin{table}
\centering
\small
\begin{threeparttable}
\caption{\label{tab:misato1k} 
Quantitative comparison on MISATO@1K.}
\renewcommand\tabcolsep{1.5pt}
\begin{tabular*}{\linewidth}{@{\extracolsep{\fill}}lccccc}
\toprule
Decoder & LPIPS$\downarrow$  & FID$\downarrow$ & U-IDS$\uparrow$ & P-IDS$\uparrow$ & G@e$\downarrow$ \\
\midrule
\midrule
SD              & 0.273 & 53.773 & 0.000 & 0.000 & 51.564 \\
ASUKA-I-SD      & \textbf{0.186} & \textbf{22.446} & \textbf{0.199}  & \textbf{0.096} & \textbf{24.791} \\
\midrule
FLUX            & 0.165 & 15.106 & 0.245  & 0.128 & 48.871 \\
PixelHacker & 0.167 & 18.515 & 0.246  & 0.139 & 39.517 \\
OmniPaint       & 0.214 & 22.009 & 0.146  & 0.061 & 73.459 \\
OmniEraser      & 0.212 & 21.669 & 0.121  & 0.041 & 82.788 \\
ASUKA-II-FLUX   & \textbf{0.156} & \textbf{14.247} & \textbf{0.265}  & \textbf{0.130} & \textbf{34.036} \\
\bottomrule
\end{tabular*}
\end{threeparttable}
\end{table}

\begin{table}
\centering
\small
\begin{threeparttable}
\caption{Number of object hallucination cases on MISATO@512. 
}
\label{tab:vlm_eval}
\renewcommand\tabcolsep{1.5pt}
\small{
\begin{tabular*}{\linewidth}{@{\extracolsep{\fill}}lcc}
\toprule 
Method & VLM Judgment $\downarrow$ & Human Judgment $\downarrow$ \\
\midrule
\midrule
SD-text & 675 & 274\\
ASUKA-I-SD & \textbf{136} & \textbf{87} \\
\midrule
FLUX-Fill & 72 & 27 \\
ASUKA-I-FLUX & 60 & 23\\
ASUKA-II-FLUX & \textbf{50} & \textbf{5} \\
\bottomrule
\end{tabular*}
}
\end{threeparttable}
\end{table}

\paragraph{Competitors}
We use the SD v1.5 inpainting model~\cite{Rombach_2022_CVPR} to analyze and compare ASUKA with competitors for a fair comparison, while validating ASUKA's generalization ability with FLUX.
We consider three SD v1.5 inpainting variants:
SD: uses a null-prompt for unconditional generation;
SD-text: uses "background" as a prompt since no captions are used in inpainting;
SD-token~\cite{wang2024repositioning}: uses learnable tokens trained with ASUKA's pipeline.
To test other ways of incorporating the MAE condition, we implement the following:
SD-IP, uses IP-Adapter~\cite{ye2023ip-adapter};
SD-T2I, uses T2I-Adapter~\cite{mou2024t2i};
SD-CAEv2, uses a CLIP-style alignment module CAEv2~\cite{zhang2023cae};
We also test SD-LaMa, which inputs LaMa~\cite{suvorov2022resolution} inpainting results instead of MAE.
We also compare with leading inpainting algorithms Co-Mod~\cite{zhao2021large}, MAT~\cite{li2022mat}, LaMa~\cite{suvorov2022resolution}, MAE-FAR~\cite{cao2022learning}, and SD-Repaint~\cite{lugmayr2022repaint},
and recent SOTA methods PixelHacker~\cite{xu2025pixelhacker}, OmniPaint~\cite{yu2025omnipaint}, and OmniEraser~\cite{wei2025omnieraser}.

\subsection{Comparison on Benchmarks}

\paragraph{Quantitative comparison} Results on Places 2 and MISATO@512 are in Tab.~\ref{tab:quantitative} 
Results of MISATO@1K are in Tab.~\ref{tab:misato1k}
Evaluation results of object hallucination are in Tab.~\ref{tab:vlm_eval}.

\emph{Quantitative comparison on SD-based models} demonstrate that:
\textbf{(1)}: Although ASUKA-I-SD is based on a fixed SD model, it consistently outperforms SD across all evaluation metrics, achieving state-of-the-art results in FID, U-IDS, and P-IDS. Notably, U-IDS and P-IDS are closely aligned with human preferences~\cite{zhao2021large} and have a potential maximum score of 0.5, highlighting ASUKA's strong performance.
\textbf{(2)}: Compared to other adapters that align the MAE prior with SD, ASUKA-I-SD shows consistently superior performance across all metrics. This demonstrates the effectiveness of our straightforward alignment module.
\textbf{(3)}: While the LaMa condition improves inpainting quality, as shown by FID and IDS scores, it is less effective than the MAE condition. When using the MAE condition as a prior, improvements can be attributed to better mitigation of object hallucination.
\textbf{(4)}: The LPIPS scores indicates that  with a frozen SD, ASUKA achieves consistent improvements but the margin will remain constrained by the frozen U-Net.
\textbf{(5)}: The G@e metric suggests that ASUKA-I-SD shows significant improvements over all SD variants, highlighting its enhanced color consistency.
These results confirm that ASUKA-I-SD improves color consistency and mitigation of object hallucination in inpainting, even when using frozen latent inpainting models. This advantage is evident both in the in-distribution dataset Places2 and the out-of-distribution dataset MISATO, and is generalizable to high-resolution image inpainting (see Tab.~\ref{tab:misato1k}).

\emph{Quantitative comparisons on FLUX-based models} show that compared with the standard FLUX-Fill model, ASUKA-II-FLUX consistently improves all evaluation metrics across both benchmark datasets and multiple resolutions (see Tabs.~\ref{tab:quantitative} and \ref{tab:misato1k}), demonstrating the generality of the proposed ASUKA framework.
In particular, the color consistency is substantially improved, as evidenced by the G@e metric.

\emph{Quantitative comparison on object hallucination} are reported in Tab.~\ref{tab:vlm_eval}.
For SD-based models, the proposed ASUKA framework substantially reduces object hallucination.
For FLUX-Fill, due to its improved model capacity and stronger condition coherence, object hallucination is largely mitigated; however, it still occurs in some seemingly simple cases (as seen in Fig~\ref{fig:flux-visual}).
We report both VLM-based judgments and human judgments to examine the consistency between automatic and human evaluations.
The results indicate that:
(1) ASUKA-I alleviates object hallucination, but the improvement is limited;
(2) ASUKA-II, in contrast, significantly reduces object hallucination;
(3) although the VLM evaluator is stricter than human evaluators, the relative ranking of different models remains largely consistent.

\paragraph{Qualitative comparison}
Fig.~\ref{fig:main-result},~\ref{fig:flux-visual} and~\ref{fig:comparison} 
show SD- and FLUX-based results.
GAN and deterministic inpainting baselines often produce blurry or boundary-inconsistent completions. 
SD variants tend to hallucinate context-unsupported objects and suffer from color shifts.
Recent SOTA methods still hallucinate
objects or leave visible artifacts.
ASUKA-I-SD reduces these failures for U-Net diffusion models.
For FLUX, ASUKA-I provides partial control, whereas ASUKA-II more consistently suppresses spurious objects through per-layer MAE injection and scaled positional alignment.

\subsection{Further Analysis of ASUKA}
\label{sec:ablation}

\paragraph{Ablation study}
We ablate the contribution of ASUKA-I-SD's two main modules in Tab.~\ref{tab:effect-module} and ASUKA-II-FLUX's modules in Tab.~\ref{tab:flux-ablation}.
Detailed decoder (Tab.~\ref{tab:decoder-ablation}), 
alignment (Tab.~\ref{tab:align-ablation}), 
$p$-schedule (Tab.~\ref{tab:ablate-p}), 
and MAE-prior (Tab.~\ref{tab:pt-vs-ft}) ablations, 
statistical (Tab.~\ref{tab:misato512_seedvar}) and efficiency (Tab.~\ref{tab:misato512_cost}) analysis,
and more comparison with text-guided inpainting (Tab.~\ref{tab:asuka-vs-text}), text-guided-decoder (Tab.~\ref{tab:decoder-t2i}), and additional datasets (Tab.~\ref{tab:celeba-ffhq}) are provided in Supplementary Sec.~\ref{sec:supp-ablations}.

\begin{table}
\centering
\caption{Effect of each module in ASUKA-I-SD.\label{tab:effect-module}}
\small
\renewcommand\tabcolsep{1.5pt}
\begin{tabular*}{\linewidth}{@{\extracolsep{\fill}}lccccc}
\toprule
Model & LPIPS$\downarrow$ & FID$\downarrow$ & U-IDS$\uparrow$ & P-IDS$\uparrow$ & G@e$\downarrow$ \\
\midrule
\midrule
SD w/ MAE      & 0.157 & 12.093 & 0.397 & 0.236 & 62.845 \\
SD w/ decoder  & 0.159 & 12.075 & 0.411 & 0.283 & 49.376 \\
ASUKA-I-SD     & \textbf{0.150} & \textbf{11.495} & \textbf{0.423} & \textbf{0.312} & \textbf{47.753} \\
\bottomrule
\end{tabular*}
\end{table}

\newcommand{\cmark}{\ensuremath{\checkmark}}
\newcommand{\xmark}{\ensuremath{-}}

\begin{table}[t]
\centering
\small
\setlength{\tabcolsep}{5pt}
\caption{\label{tab:flux-ablation}
Ablation of the components in ASUKA-II-FLUX.}
\begin{tabular}{@{}lccccc@{}}
\toprule
Variant& LPIPS$\downarrow$ & FID$\downarrow$ & U-IDS$\uparrow$ & P-IDS$\uparrow$ & G@e$\downarrow$ \\
\midrule
\midrule
MAE LoRA& 0.143 & 11.063 & 0.382 & 0.213 & 51.080 \\
 + new PE& \textbf{0.139} & 10.997 & 0.381 & 0.215 & 51.992 \\
\quad + task prompt & 0.142 & 10.975 & 0.382 & 0.217 & 50.866 \\
\quad \quad + gate       & 0.140 & 10.772 & \textbf{0.383} & 0.236 & 49.724 \\
\midrule
\quad \quad \quad + decoder & \textbf{0.139} & \textbf{10.681} & \textbf{0.383} & \textbf{0.241} & \textbf{48.793} \\
\bottomrule
\end{tabular}
\end{table}

\paragraph{Human preference}
To complement automatic metrics, we conduct a user study on 40 randomly selected testing images with 100 valid anonymous responses.
Participants choose the best result for unwanted-object mitigation (UOM) and color consistency (CC).
ASUKA achieves the highest preference for both SD-based and FLUX-based comparisons, as  in Tab.~\ref{tab:user-study-compact}.

\begin{table}[t]
\centering
\caption{Human preference: top-1 ratio.
See full SD results in 
Tab.~\ref{tab:user-study}.
\label{tab:user-study-compact}}
\small
\renewcommand\tabcolsep{2.0pt}
\begin{tabular*}{\linewidth}{@{\extracolsep{\fill}}lcc}
\toprule
Model group / method & UOM (\%) & CC (\%) \\
\midrule
\midrule
Best non-ASUKA SD-based baseline & 16.18 & 15.83 \\
ASUKA-I-SD & \textbf{39.43} & \textbf{40.05} \\
\midrule
FLUX-Fill & 18.79 & 22.79 \\
ASUKA-I-FLUX & 32.88 & 33.09 \\
ASUKA-II-FLUX & \textbf{48.32} & \textbf{44.12} \\
\bottomrule
\end{tabular*}
\end{table}

\section{Conclusion}
In this paper, we proposed Aligned Stable inpainting with Unknown Areas prior (ASUKA) to achieve unwanted-object-mitigated and color-consistent inpainting via frozen latent inpainting models.
To avoid unwanted object insertion, we adopt a reconstruction-based masked auto-encoder (MAE) as the context-stable prior for masked region purely from unmasked region.
Then we align the context-stable prior to frozen generative models with the proposed alignment module to inject the MAE priors in a cross-attention manner for both U-Net and Transformer based models to ensure the effectiveness of guidance.
To achieve color-consistency, we resolve the mask-unmask color inconsistency in the latent decoding process.
We train an unmask-region conditioned VAE decoder to perform local harmonization during the decoding process.
To validate the efficacy of inpainting algorithms in different image domains and mask types, we introduce an evaluation dataset, named as MISATO, from existing datasets.
We propose two new metrics to explicitly evaluate the object hallucination and color-consistency of inpainted images.
ASUKA enjoys unwanted-object-mitigated and color-consistent inpainting results and superior than leading inpainting models.


\bibliographystyle{IEEEtran}
\bibliography{main}

\ifdefined\ASUKAWithSupplement
  \clearpage
  \onecolumn
  \def\ASUKASupplementaryInput{}
  \ifdefined\ASUKASupplementaryInput
  \def\ASUKAEndSupplementary{}
\else
  \documentclass[lettersize,onecolumn]{IEEEtran}
  \usepackage{amsmath,amsfonts,amssymb}
  \usepackage{graphicx}
  \usepackage{booktabs}
  \usepackage{multirow}
  \usepackage{array}
  \usepackage{cite}
  \usepackage{xcolor}
  \usepackage{colortbl}
  \usepackage{threeparttable}
  \usepackage{hyperref}
  \usepackage{makecell}
  \usepackage{bm}
  \def\ASUKAEndSupplementary{\end{document}}
\fi

\ifdefined\ASUKASupplementaryInput
\else
  \definecolor{rank1}{RGB}{226, 164, 145}
  \definecolor{rank2}{RGB}{235, 197, 185}
  \definecolor{rank3}{RGB}{244, 227, 222}
\fi

\setcounter{section}{0}
\setcounter{figure}{0}
\setcounter{table}{0}
\setcounter{equation}{0}
\renewcommand{\thesection}{S\arabic{section}}
\renewcommand{\thefigure}{S\arabic{figure}}
\renewcommand{\thetable}{S\arabic{table}}
\renewcommand{\theequation}{S\arabic{equation}}
\renewcommand{\theHsection}{suppsection.\arabic{section}}
\renewcommand{\theHfigure}{suppfigure.\arabic{figure}}
\renewcommand{\theHtable}{supptable.\arabic{table}}
\renewcommand{\theHequation}{suppequation.\arabic{equation}}

\ifdefined\ASUKASupplementaryInput
  \section*{Supplementary Material}
\else
  \begin{document}

  \title{Supplementary Material for ``Aligned Stable Inpainting: Mitigating Unwanted Object Insertion and Preserving Color Consistency''}
  \author{Yikai Wang*, Junqiu Yu*, Chenjie Cao, Xiangyang Xue, Yanwei Fu}
  \maketitle
\fi

This supplementary file contains the additional explanation and analysis, more experimental results and visualization for our manuscript.
Table~\ref{tab:supp-map} shows the relationship between the content and the corresponding supplementary and manuscript  section.

\begin{table}[t]
\centering
\caption{Mapping between the content, supplementary section, and the manuscript section.}
\label{tab:supp-map}
\small
\begin{tabular*}{\linewidth}{@{\extracolsep{\fill}}lll}
\toprule
Content & Supplementary section & Manuscript section \\
\midrule
\midrule
Formal Problem Definition & Sec.~\ref{sec:supp-problem-definition} 
& Problem setup 
(Sec.~\ref{sec:problem-setup})
\\
Further Discussion: Empirical Perspective & Sec.~\ref{sec:supp-empirical-discuss} & 
Problem setup 
(Sec.~\ref{sec:problem-setup})
\\
Further Discussion: Theoretical Perspective &  Sec.~\ref{sec:supp-theoretical-discuss} & Theoretical Perspective: Learned Posterior Regularization 
(Sec.~\ref{sec:analysis})
\\
MISATO Dataset & Sec.~\ref{sec:supp-dataset} & Evaluation datasets 
(Sec.~\ref{sec:dataset})
\\
Additional Ablations and Extended Results & Sec.~\ref{sec:supp-ablations} & Further Analysis of ASUKA 
(Sec.~\ref{sec:ablation})
\\
\bottomrule
\end{tabular*}
\end{table}

\section{Formal Problem Definition}
\label{sec:supp-problem-definition}
\paragraph{Inpainting setup}
Let $\bm{x}\in[0,1]^{H\times W\times 3}$ be an image and $M\in\{0,1\}^{H\times W}$ be a binary mask, where $M_i=1$ indicates the unknown region.
The visible observation is
\begin{equation}
\bm{x}_{\bar M}=(1-M)\odot \bm{x}.
\end{equation}
An inpainting model generates a completion
\begin{equation}
\hat{\bm{x}}=\bm{x}_{\bar M}+M\odot G_{\theta}(\bm{x}_{\bar M},M,c,\epsilon),
\end{equation}
where $c$ denotes optional user guidance and $\epsilon$ is the random noise.

\paragraph{Unwanted object insertion}
Unwanted object insertion is a task-conditioned semantic failure.
It cannot be defined purely by pixel error, because the ground-truth masked region may itself contain an object that should not be regenerated.
Let $\mathcal{A}(\hat{\bm{x}},M)$ denote the set of confident object-like semantic instances detected inside the generated masked region.
Each element $a\in\mathcal{A}$ represents an object instance with category, location, and visual attributes.
For a task $\mathcal{T}$, define the context-supported semantic set as
\begin{equation}
\mathcal{S}_{\tau}(\bm{x}_{\bar M},M,c)=
\{a: p_{\mathcal{T}}(a\mid \bm{x}_{\bar M},M,c)\ge \tau\},
\end{equation}
where $p_{\mathcal{T}}(a\mid \bm{x}_{\bar M},M,c)$ is the probability that semantic instance $a$ is supported by the visible context and task condition.
For example, an object explicitly requested by a prompt or strongly implied by visible structure belongs to $\mathcal{S}_{\tau}$, while an arbitrary foreground object generated in a background mask does not.
We define the unwanted object insertion indicator as
\begin{equation}
\mathrm{UOI}_{\tau}(\hat{\bm{x}};\bm{x}_{\bar M},M,c)
=
\mathbb{I}\left[
\exists a\in \mathcal{A}(\hat{\bm{x}},M)\ \mathrm{s.t.}\ a\notin \mathcal{S}_{\tau}(\bm{x}_{\bar M},M,c)
\right].
\end{equation}
The hallucination rate of an inpainting model is
\begin{equation}
\mathrm{HR}(G_{\theta})
=
\mathbb{E}_{(\bm{x},M,c),\epsilon}
\left[
\mathrm{UOI}_{\tau}(\hat{\bm{x}};\bm{x}_{\bar M},M,c)
\right].
\end{equation}
In practice, $\mathcal{A}$ and $\mathcal{S}_{\tau}$ are approximated by human annotation or VLM-based judgment.

\paragraph{Color inconsistency}
Color inconsistency is a low-frequency discontinuity between the generated masked region and the visible unmasked region.
Let $L_{\sigma}(\cdot)$ be a low-pass filtering operator, optionally applied in a perceptual color space such as Lab.
Let $B_r(M)$ denote an $r$-pixel band around the mask boundary.
We define the boundary color inconsistency with respect to the ground-truth image as
\begin{equation}
\mathrm{CI}_{r}(\hat{\bm{x}};\bm{x},M)
=
\frac{1}{|B_r(M)|}
\sum_{i\in B_r(M)}
\left\|
\nabla L_{\sigma}(\hat{\bm{x}})_i-\nabla L_{\sigma}(\bm{x})_i
\right\|_1.
\end{equation}
A result is color-inconsistent if $\mathrm{CI}_{r}(\hat{\bm{x}};\bm{x},M)>\delta_c$, where $\delta_c$ is a tolerance threshold.
This definition focuses on low-frequency boundary mismatch rather than exact texture reconstruction.

\paragraph{Decoder-level color shift}
For latent inpainting, it is also useful to define the color shift before compositing.
Let $\tilde{\bm{x}}=D(\hat{\bm{z}})$ be the raw decoded image.
The decoder-induced visible-region color bias is
\begin{equation}
\mathrm{CS}_{D}
=
\frac{1}{|\bar M|}
\left\|
(1-M)\odot
\left(
L_{\sigma}(\tilde{\bm{x}})-L_{\sigma}(\bm{x})
\right)
\right\|_1.
\end{equation}
A large $\mathrm{CS}_{D}$ indicates that the decoder produces a global or local color shift even where ground-truth visible pixels are known.
After compositing, this bias becomes a visible seam between generated and unmasked regions.

\section{Further Discussion: Empirical Perspective}
\label{sec:supp-empirical-discuss}
In this part, we discuss the definitions and root causes of two issues we studied from the empirical perspective.
\subsection{Unwanted Object Insertion}
\paragraph{Definition}
In generative inpainting, the output is expected to be guided by both the unmasked region of the image and user-provided condition.
Ideally, the generated content should remain consistent with these sources of guidance.
In practice, however, models sometimes introduce random or irrelevant elements into the masked region.
These elements may be meaningful objects or meaningless artifacts. We refer to this issue as unwanted object insertion since they cannot be estimated directly from the surrounding context.

\paragraph{Root Cause}
The occurrence of unwanted object insertion can be traced to the interplay of three fundamental components in generative inpainting:
the initial noise that establishes the starting state of the generation process,
the unmasked region that communicates information to the masked region through spatial-aware self-attention mechanisms,
and the user guidance that interacts with the model through cross-attention or extended self-attention layers (as in MMDiT).
When any of these components fail to function as intended, the model may generate objects that do not align with either the visual context or the user’s guidance.

\paragraph{Hallucination from initial noise}
Although diffusion and flow models are designed to connect real image distributions with random Gaussian noise, inpainting models are typically fine-tuned on large datasets using random masking strategies with global captions describing the overall image.
In certain cases, entire objects may be masked while absent from the caption, yet the model is trained to reconstruct them.
This introduces a contradiction: the model may incorrectly infer that particular noise patterns carry semantic meaning and, as a result, generate objects irrespective of the surrounding context or user guidance.
To address this issue, a fine-grained training strategy that eliminates inconsistencies between masking and captions would be ideal.
As an alternative, post-training methods can be employed to strengthen the influence of other guidance signals, thereby reducing reliance on spurious semantic noise.

\paragraph{Hallucination from the unmasked region}
The visible portion of the image may inadvertently reveal information about the masked area, leading the model to generate objects that should not appear.
This problem can arise when shadows of a removed object remain visible, prompting the model to regenerate the object for the sake of realism.
Similarly, the shape of a mask may `betray' the nature of the object that was removed, biasing the model toward recreating a similar content.
In images containing multiple instances of the same object, masking only one can also inspire the model to reproduce it, influenced by the presence of the others.
More subtly, high-frequency cues in the surrounding texture may leak information that encourages the reconstruction of unintended content.
These challenges can be mitigated in several ways: the model may be trained to disregard shadows or to remove them alongside the masked object as part of a harmonization process; mask shapes can be perturbed to reduce semantic informativeness; and morphological dilation of the mask, typically by five to twenty pixels, can eliminate residual details that might otherwise bias generation.
More generally, improvements in the model’s ability to follow user prompts are essential to ensuring that use guidance take precedence over unmasked-region cues.

\paragraph{Inability to follow guidance}
In this case, the problem does not lie in the prompt itself but rather in the model’s limited capacity to integrate and prioritize it.
In attention–based architectures, residual connections sometimes weaken the influence of prompts, so altering the guidance may have little impact on the final output.
Increasing the weighting of conditioning features can amplify the effect of guidance, though this often comes at the cost of reduced visual fidelity.
A more effective strategy is to adjust the prompt-injection mechanism by modifying the layer inputs directly.

In sum, unwanted object insertion arises from the interaction of semantic noise, residual signals in the unmasked region, and limitations in prompt integration. Addressing these issues requires solutions at both the training and inference stages, ranging from improved data strategies and architectural adjustments to post-processing interventions. Such measures are crucial for enhancing both the controllability and reliability of generative inpainting models.
In this paper, we propose a post-training method to improve existing pre-trained generative inpainting models. Our approach introduces a context-stable prior, automatically derived from the unmasked region, to guide the inpainting process.

\subsection{Color Inconsistency}
\paragraph{Color-inconsistency is a general problem}
In generative inpainting models, there is often a mismatch in color between the masked and unmasked regions.
This happens when the generated (masked) area shows a color shift compared to the surrounding unmasked area.
As shown in 
Fig.~\ref{fig:color-shift}, 
such shifts can occur in many situations, including indoor and outdoor scenes, random or continuous masks, and may appear as either darker or lighter regions.
These shifts are mainly caused by imperfections in the VAE and the latent generator.

\paragraph{Information loss of VAE}
Popular latent diffusion and rectified flow models generate images entirely in the latent space and then decode these latent codes back into image space using a VAE.
Although the decoder is trained to reconstruct the original image, it still suffers from information loss.
This problem is especially evident in tasks like inpainting, where the unmasked region has ground-truth values.

Rombach \emph{et. al.}~\cite{Rombach_2022_CVPR} argued that the diffusion model should focus on semantic compression, while the VAE handles perceptual compression with high-frequency details.
However, we show that \emph{low-frequency reconstruction loss in the VAE cannot be ignored}, as illustrated in 
Fig.~\ref{fig:info_loss} 
(b).
The VAE not only degrades high-frequency details but also introduces noticeable color shifts.
This effect becomes clear when the VAE is applied repeatedly, as shown in 
Fig.~\ref{fig:info_loss} 
(a), where the color shift grows stronger with each reconstruction.
Since humans are highly sensitive to changes in low-frequency image information, even subtle color shifts can create significant inconsistencies.
The issue becomes worse  with irregular or large masks.

\paragraph{Gap between real and generated latents}
Besides the information loss caused by the VAE during reconstruction, there is also a gap between the generated latents and the real ones.
This gap also leads to color inconsistency, even when the VAE reconstruction loss is reduced, as shown in 
Fig.~\ref{fig:latent-aug-visualization}.
To achieve better color consistency, we need to address both the VAE’s reconstruction loss and the latent generator.

\section{Further Discussion: Theoretical Perspective}
\label{sec:supp-theoretical-discuss}

\paragraph{Inpainting as an ill-posed inverse problem}
Image inpainting is intrinsically ill-posed: the visible region $\bm{x}_{\bar M}$ and the mask $M$ do not uniquely determine the missing content $\bm{x}_M$.
Many completions can be visually plausible, and a generative model therefore samples from a conditional distribution
\begin{equation}
p_{\theta}(\bm{z}_M\mid \bm{z}_{\bar M},M,c),
\end{equation}
where $\bm{z}=E(\bm{x})$ is the latent code and $c$ denotes the conditioning signal.
If this conditional distribution contains modes that are plausible under the global image prior but weakly supported by the visible context, sampling may produce unwanted object insertion.
From the inverse-problem perspective~\cite{engl2014inverse}, suppressing such modes requires an additional regularizer that favors context-supported completions.

\paragraph{MAE as a conditional-mean prior}
Let $Y=(\bm{x}_{\bar M},M)$ denote the visible observation.
A masked autoencoder trained with an $\ell_2$ reconstruction loss solves
\begin{equation}
r_{\phi}^{*}
=
\arg\min_r
\mathbb{E}
\left[
\|\bm{x}_M-r(Y)\|_2^2
\right].
\end{equation}
By the Bayes estimator property under squared loss, the optimal solution is
\begin{equation}
r_{\phi}^{*}(Y)=
\mathbb{E}[\bm{x}_M\mid\bm{x}_{\bar M},M].
\end{equation}
Thus, the MAE prediction is not an arbitrary reconstruction heuristic.
It estimates the conditional mean of the missing region given the visible context.
This explains two empirical properties: the MAE prior is stable because unsupported semantic modes are averaged out unless strongly implied by the visible region, and the MAE output is often blurry because multiple valid completions are collapsed into a conditional mean.
This is why ASUKA uses MAE as a stabilizing prior for a high-fidelity latent generator rather than directly using MAE as the final output.
This interpretation is consistent with denoising autoencoder theory, where reconstruction residuals point toward high-density regions of the data distribution~\cite{vincent2011connection,alain2014regularized}.

\paragraph{Posterior regularization view}
Let $\bm{m}_{\phi}=r_{\phi}(\bm{x}_{\bar M},M)$ be the MAE reconstruction prior.
Define a soft consistency energy
\begin{equation}
d(\bm{z}_M,\bm{m}_{\phi})^2
=
\left\|
P_M D(\bm{z})-\bm{m}_{\phi}
\right\|_2^2,
\end{equation}
where $D$ is the decoder and $P_M$ selects the masked region.
ASUKA can be interpreted as sampling from a regularized posterior distribution:
\begin{equation}
p_{\theta,\lambda}(\bm{z}_M\mid\bm{z}_{\bar M},M,\bm{m}_{\phi})
=
\frac{
p_{\theta}(\bm{z}_M\mid\bm{z}_{\bar M},M)
\exp[-\lambda d(\bm{z}_M,\bm{m}_{\phi})^2]
}{Z_{\lambda}}.
\label{eq:supp-regularized}
\end{equation}
This is a product of the original generative prior and a learned context-consistency likelihood.
It is analogous to posterior sampling in diffusion inverse problems, where the reverse process combines a generative score with a data-consistency or likelihood term~\cite{song2021scorebased,chung2023diffusion,kawar2022denoising,lugmayr2022repaint}.
Here, the MAE prior plays the role of a learned pseudo-measurement of what the visible context supports.

\paragraph{Hallucination suppression}
Consider completions close to the MAE prior,
$\mathcal{A}_r=\{\bm{z}_M:d(\bm{z}_M,\bm{m}_{\phi})\le r\}$,
and far completions,
$\mathcal{H}_{\tau}=\{\bm{z}_M:d(\bm{z}_M,\bm{m}_{\phi})\ge \tau\}$, with $\tau>r$.
Under Eq.~\eqref{eq:supp-regularized},
\begin{equation}
\frac{
p_{\theta,\lambda}(\mathcal{H}_{\tau})
}{
p_{\theta,\lambda}(\mathcal{A}_r)
}
\le
\exp[-\lambda(\tau^2-r^2)]
\frac{
p_{\theta}(\mathcal{H}_{\tau})
}{
p_{\theta}(\mathcal{A}_r)
}.
\end{equation}
Therefore, modes far from the MAE prior are exponentially down-weighted relative to modes close to the context-supported prior.
If unwanted inserted objects correspond to completions that strongly deviate from the conditional-mean background estimate, ASUKA suppresses them by construction.

\paragraph{Diversity-stability trade-off}
For the regularized distribution in Eq.~\eqref{eq:supp-regularized},
\begin{equation}
\frac{d}{d\lambda}
\mathbb{E}_{p_{\theta,\lambda}}
\left[
d(\bm{z}_M,\bm{m}_{\phi})^2
\right]
=
-
\mathrm{Var}_{p_{\theta,\lambda}}
\left[
d(\bm{z}_M,\bm{m}_{\phi})^2
\right]
\le 0.
\end{equation}
Increasing the strength of the MAE prior monotonically concentrates samples around the context-supported estimate.
This improves hallucination suppression, but excessive regularization would eventually make outputs overly conservative.
ASUKA avoids this degenerate regime by using the MAE prior as an aligned condition for a frozen generator, rather than directly replacing the generated region with the MAE reconstruction.
The generator therefore retains its learned natural-image distribution and high-frequency synthesis ability.
In this sense, ASUKA suppresses context-unsupported semantic diversity, such as arbitrary foreground object insertion, while preserving context-supported appearance diversity, including textures, local structures, and illumination variations derivable from the surrounding scene.

\paragraph{Alignment as amortized guidance}
Directly sampling from Eq.~\eqref{eq:supp-regularized} would require iterative optimization or gradient guidance through the decoder.
ASUKA instead amortizes this guidance into a trainable alignment module
\begin{equation}
C_{\mathrm{MAE}}=A_{\psi}(F_{\mathrm{MAE}}),
\end{equation}
while keeping the generator frozen.
For diffusion models, the alignment module is optimized through the standard denoising objective
\begin{equation}
\min_{\psi}
\mathbb{E}
\left[
\left\|
\epsilon-
\epsilon_{\theta}
(
\bm{z}_t,t,\bm{z}_{\bar M},M,A_{\psi}(F_{\mathrm{MAE}})
)
\right\|_2^2
\right],
\end{equation}
and for rectified-flow models through the corresponding velocity prediction objective.
Since diffusion denoising objectives estimate conditional score fields~\cite{ho2020denoising,song2021scorebased}, training $A_{\psi}$ learns a condition that steers the frozen score field toward MAE-consistent completions.

\paragraph{Why ASUKA-II uses gated per-layer injection}
In MMDiT-style generators, text tokens are repeatedly transformed together with image tokens.
Directly replacing text tokens with MAE features forces the frozen model to process an out-of-distribution condition through a deep text pathway.
ASUKA-II instead injects the MAE prior as a gated residual condition at each layer.
The task prompt keeps the conditioning stream close to the pretrained model's condition manifold, while the MAE branch adds a learned correction corresponding to the context-stability regularizer.
The gate controls the local regularization strength, preventing the MAE prior from overwhelming the generator.
Scaled positional encoding ensures that this regularizer is applied to the correct spatial location.

\paragraph{Color-consistent decoding as conditional denoising}
A standard VAE decoder estimates pixels from latent codes alone, but in inpainting the final image combines generated masked pixels with real unmasked pixels.
Low-frequency errors therefore become visible as color discontinuities.
ASUKA trains a conditional decoder
\begin{equation}
\hat{\bm{x}}=D_{\eta}(\hat{\bm{z}},\bm{x}_{\bar M},M)
\end{equation}
with color and latent augmentations.
Under an $\ell_2$ reconstruction objective, the optimal decoder estimates
\begin{equation}
D_{\eta}^{*}
=
\mathbb{E}
\left[
\bm{x}\mid\hat{\bm{z}},\bm{x}_{\bar M},M
\right].
\end{equation}
Thus, the decoder learns a conditional denoising projection from corrupted latent/color statistics back to the clean image, using the visible region as an anchor.
In ASUKA, the corruption corresponds to VAE-induced color shift and generated-real latent mismatch, making the decoder theoretically grounded as a local harmonization estimator.

\section{MISATO Dataset}
\label{sec:supp-dataset}

\begin{figure}[h]
\centering
\includegraphics[width=0.75\linewidth]{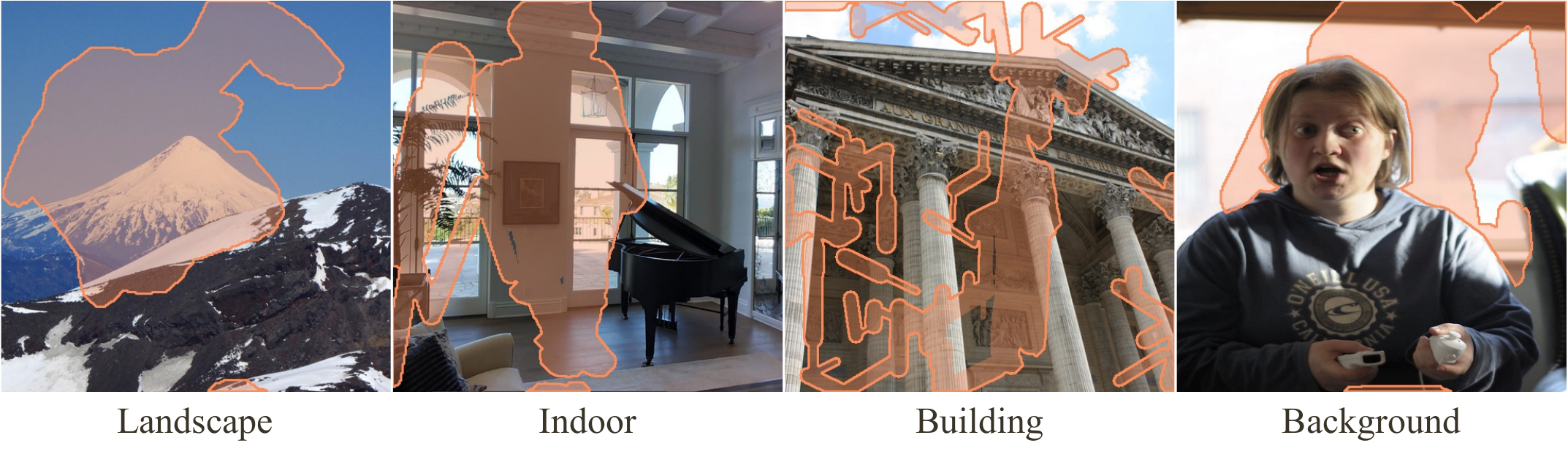}
\caption{Different image domains in MISATO, covering indoor, landscape, building, and background inpainting.}
\label{fig:supp-misato}
\end{figure}

MISATO is built from Matterport3D~\cite{Matterport3D}, Flickr-Landscape~\cite{lin2021infinitygan}, MegaDepth~\cite{MDLi18}, and COCO 2014~\cite{lin2014microsoft}.
The principle of constructing MISATO is to select the most representative and diverse examples.
To this end, for first three datasets, we use CLIP visual model~\cite{radford2021learning} to extract semantic visual features. 
Then we use BisectingKMeans~\cite{steinbach2000comparison} to cluster each dataset into 500 clusters, and select the cluster centers as the evaluation data.
The selected data are center cropped and then resized to $512^2$.
For COCO, we focus on the background inpainting. 
To this end, for each data we identify the foreground with provided segmentation and remove it from the generated masks, yielding a dataset specified for purely background inpainting.
An illustration of the dataset is in Fig.~\ref{fig:supp-misato}.

\section{Additional Ablations and Extended Results}
\label{sec:supp-ablations}

\begin{table}[h]
\centering
\small
\begin{threeparttable}
\caption{\label{tab:decoder-ablation}Comparison of different decoders for SD.}
\renewcommand\tabcolsep{2.0pt}
\begin{tabular*}{\linewidth}{@{\extracolsep{\fill}}lccccc}
\toprule
Decoder & LPIPS$\downarrow$ & FID$\downarrow$ & U-IDS$\uparrow$ & P-IDS$\uparrow$ & G@e$\downarrow$ \\
\midrule
\midrule
VAE        & 0.156 & 11.949 & 0.387 & 0.253 & 63.142 \\
+ cond.    & 0.151 & 11.634 & 0.410 & 0.272 & 48.588 \\
+ color    & 0.152 & 11.603 & 0.407 & 0.273 & 49.538 \\
\midrule
Ours       & \textbf{0.150} & \textbf{11.495} & \textbf{0.423} & \textbf{0.312} & \textbf{47.753} \\
\bottomrule
\end{tabular*}
\end{threeparttable}
\end{table}

\paragraph{Ablation of decoder}
For the decoder, we compare ASUKA-SD with 
(1) \textit{VAE}: the decoder used in SD;
(2) \textit{+ cond.}: the decoder conditioned on unmasked image~\cite{zhu2023designing};
(3) \textit{+ color}: only trained with color augmentation ;
Results are in Tab.~\ref{tab:decoder-ablation}, showing the superiority of our decoder.

\begin{table}[h]
\centering
\begin{threeparttable}
\renewcommand\tabcolsep{2.0pt}
\small
\caption{\label{tab:align-ablation}Ablation of different alignment modules.}
\begin{tabular*}{\linewidth}{@{\extracolsep{\fill}}lccccc}
\toprule
Align & LPIPS$\downarrow$ & FID$\downarrow$ & U-IDS$\uparrow$ & P-IDS$\uparrow$ & G@e$\downarrow$ \\
\midrule
\midrule
linear    & 0.155 & 11.934 & 0.400 & 0.263 & 48.983 \\
attn      & 0.152 & 11.613 & 0.403 & 0.268 & 48.785 \\
cross x4  & 0.152 & 11.762 & 0.405 & 0.256 & 48.279 \\
\midrule
Ours      & \textbf{0.150} & \textbf{11.495} & \textbf{0.423} & \textbf{0.312} & \textbf{47.753} \\
\bottomrule
\end{tabular*}
\end{threeparttable}
\end{table}

\paragraph{Ablation of alignment module}
We validate the efficacy of our alignment module step by step:
(1) \textit{linear}: Use linear layer to align feature dimension only;
(2) \textit{attn}: Based on \textit{linear}, further use a single self-attention block to align the distribution;
(3) \textit{cross x4}: we instead use learnable query and 4 cross-attention layers to learn the MAE prior.
ASUKA-I-SD adopts 4 self-attention blocks.
Results are shown in Tab.~\ref{tab:align-ablation}. 
The self-attention block shows improved results compared with only align dimension and cross-attention block.
Using 4 self-attention blocks improves the capacity.

\begin{table}[h]
\centering
\renewcommand\tabcolsep{1.5pt}
\caption{Comparison of ASUKA with text-guided SD \label{tab:asuka-vs-text}}
\small
\begin{tabular*}{\linewidth}{@{\extracolsep{\fill}}lccccc}
\toprule
Model  & LPIPS$\downarrow$ & FID$\downarrow$ & U-IDS$\uparrow$ & P-IDS$\uparrow$ & G@e$\downarrow$ \\
\midrule
\midrule
SD (BLIP2)   & 0.163  & 12.536 & 0.370 & 0.225 & 70.846 \\
ASUKA-I-SD   & \textbf{0.150} & \textbf{11.495} & \textbf{0.423} & \textbf{0.312} & \textbf{47.753} \\
\bottomrule
\end{tabular*}
\end{table}

\paragraph{Comparison with text-guided inpainting}
We compare ASUKA-I-SD with text-guided SD model, as shown in Tab.~\ref{tab:asuka-vs-text}.
We run SD inpainting using text captions generated by BLIP2~\cite{li2023blip}.
ASUKA performs better, since captions describe the entire image, while MAE focuses on reconstructing only the masked region, leading to more precise guidance.

\begin{table}[h]
\centering
\caption{Ablation of $p$. \label{tab:ablate-p}}
\small
\renewcommand\tabcolsep{1.5pt}
\begin{tabular*}{\linewidth}{@{\extracolsep{\fill}}lccccc}
\toprule
Model  & LPIPS$\downarrow$ & FID$\downarrow$ & U-IDS$\uparrow$ & P-IDS$\uparrow$ & G@e$\downarrow$ \\
\midrule
\midrule
p=0            & 0.155 & 11.804 & 0.403 & 0.288 & 48.032 \\
p=1            & 0.152 & 11.734 & 0.394 & 0.296 & 47.997 \\
linear decay p & 0.152 & 11.558 & 0.405 & 0.307 & 47.814 \\
Ours           & \textbf{0.150} & \textbf{11.495} & \textbf{0.423} & \textbf{0.312} & \textbf{47.753} \\
\bottomrule
\end{tabular*}
\end{table}

\paragraph{Ablation of $p$}
We analyze how different values of $p$ affect ASUKA in Tab.~\ref{tab:ablate-p}. 
The results show that our warm-up and freeze strategy outperforms other approaches.

\begin{table}[h]
\centering
\caption{Additional results on benchmark datasets \label{tab:celeba-ffhq}}
\footnotesize
\renewcommand\tabcolsep{1.5pt}
\begin{tabular*}{\linewidth}{@{\extracolsep{\fill}}llccccc}
\toprule
Dataset & Model & LPIPS$\downarrow$ & FID$\downarrow$ & U-IDS$\uparrow$ & P-IDS$\uparrow$ & G@e$\downarrow$ \\
\midrule
\midrule
\multirow{5}{*}{CelebA-HQ} &
SD & 0.132 & 11.968 & 0.282 & 0.101 & 42.870 \\
& ASUKA-I-SD & 0.129 & 10.190 & \textbf{0.293} & \textbf{0.134} & 40.503 \\
& FLUX-Fill & 0.127 & 5.720 & 0.290 & 0.098 & 44.048 \\
& ASUKA-II-FLUX & \textbf{0.126} & \textbf{5.471} & 0.262 & 0.127 & \textbf{38.981} \\
\midrule
\multirow{5}{*}{FFHQ} &
SD & 0.139 & 2.235 & 0.371 & 0.197 & 43.529 \\
& ASUKA-I-SD & 0.131 & 2.060 & \textbf{0.386} & \textbf{0.205} & \textbf{30.848} \\
& FLUX-Fill & \textbf{0.127} & 2.310 & 0.323 & 0.118 & 46.105 \\
& ASUKA-II-FLUX & 0.128 & \textbf{1.844} & 0.362 & 0.163 & 40.935 \\
\bottomrule
\end{tabular*}
\end{table}

\paragraph{Additional Results}
We further compare ASUKA with standard SD and FLUX inpainting model on two additional datasets: CelebA-HQ~\cite{karras2018progressive} and FFHQ~\cite{karras2019style}. As shown in Tab.~\ref{tab:celeba-ffhq}, these results provide more evidence of ASUKA’s effectiveness.

\begin{table}[h]
\centering
\caption{Our Decoder in Text-Guided Inpainting.\label{tab:decoder-t2i}}
\small
\renewcommand\tabcolsep{1.5pt}
\begin{tabular*}{\linewidth}{@{\extracolsep{\fill}}lcccccc}
\toprule
Model  & LPIPS$\downarrow$  & FID$\downarrow$ & P-IDS$\uparrow$  & U-IDS$\uparrow$ & G@e$\downarrow$\\
\midrule
\midrule
SD & 0.192 & 37.208 & 0.2445 & 0.092 & 51.345 \\
SD w/ our decoder & \textbf{0.189} & \textbf{36.532} & \textbf{0.255} & \textbf{0.098} & \textbf{36.908} \\
\midrule
FLUX-Fill & 0.121 & 18.554 & 0.369 & 0.170 & 51.272  \\
FLUX w/ our decoder & \textbf{0.102} & \textbf{16.222} & \textbf{0.404} & \textbf{0.204} & \textbf{32.485} \\
\bottomrule
\end{tabular*}
\end{table}
\paragraph{Our Decoder in Text-Guided Inpainting}
To test the generalizability of our decoder, we evaluate it on text-guided inpainting tasks. 
We compare our decoder with the original decoder using 1,000 randomly sampled images from “jackyhate/text-to-image-2M”~\cite{zk_2024}. 
The results in Tab.~\ref{tab:decoder-t2i} confirm its effectiveness for general inpainting tasks.

\begin{table}[h]
\centering
\caption{Comparison of ASUKA-I-SD using pre-trained MAE v.s. fine-tuned MAE.\label{tab:pt-vs-ft}}
\begin{tabular*}{\linewidth}{@{\extracolsep{\fill}}lcccc}
\toprule
MAE  & LPIPS$\downarrow$  & FID$\downarrow$ & U-IDS$\uparrow$ & P-IDS$\uparrow$\\
\midrule
\midrule
pre-trained & 0.151 & 11.513 & 0.354 & \textbf{0.258}\\
fine-tuned  & \textbf{0.150} & \textbf{11.460} & \textbf{0.368} & 0.256\\
\bottomrule
\end{tabular*}
\end{table}

\paragraph{Ablation of MAE prior}
We compare our fine-tuned MAE with directly adopting the MAE trained in~\cite{cao2022learning}.
To this end, we train ASUKA with the MAE in~\cite{cao2022learning} using the same training strategy and compare the results in Tab.~\ref{tab:pt-vs-ft}.
Results suggest the improvements of fine-tuning MAE, especially on FID and U-IDS.
This improvement comes from the better adaptation on the real-world masks.

\begin{figure}[h]
\centering
\includegraphics[width=0.85\linewidth]{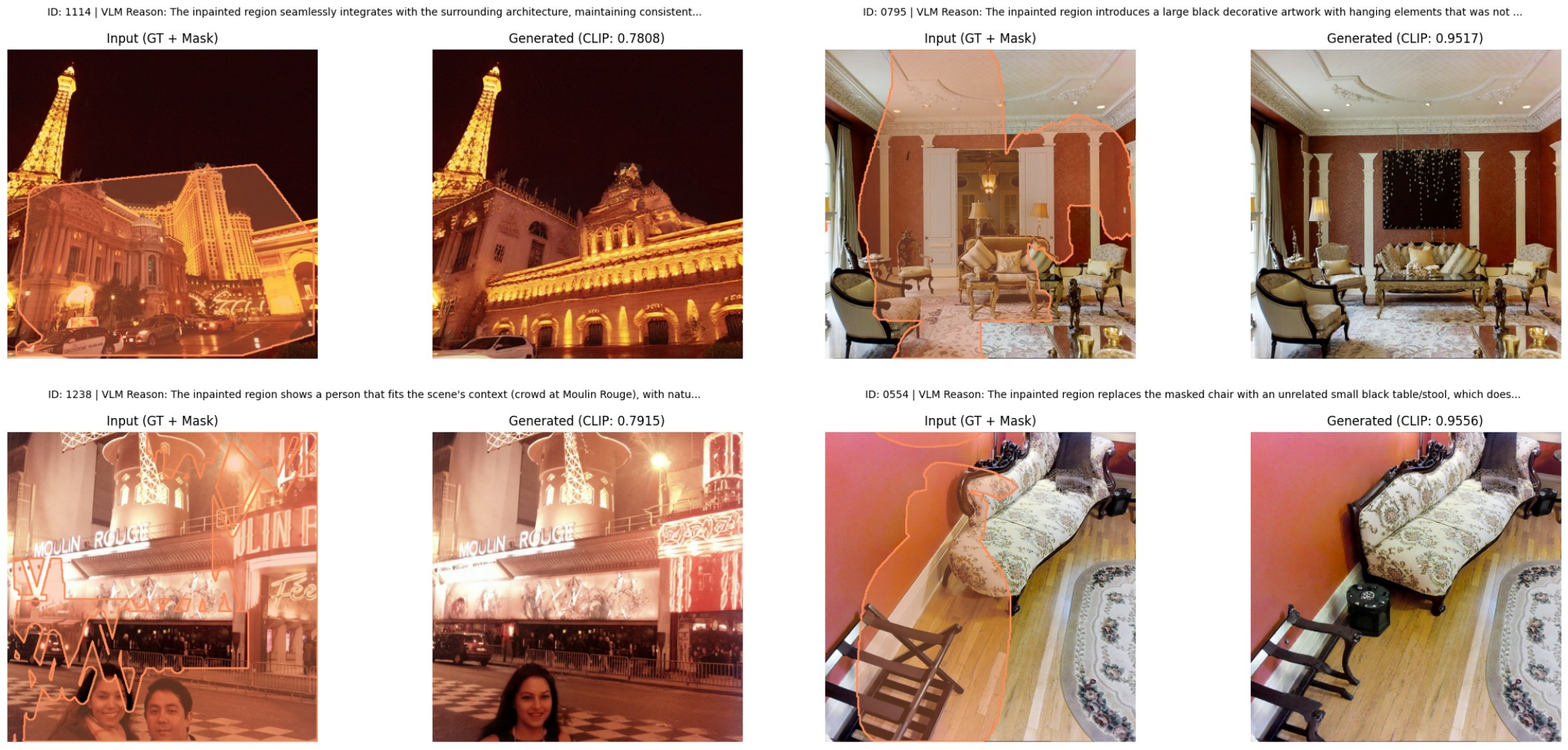}
\caption{Object hallucination evaluation using the VLM-based metric and CLIP@mask. CLIP may assign high similarity to semantically incorrect cases, while the VLM-based metric more directly judges unwanted object insertion.}
\label{fig:supp-vlm-vs-clip}
\end{figure}
\paragraph{Human and VLM Evaluation Analysis}
\label{sec:supp-vlm-clip}
In the CVPR version, we used CLIP@mask, which computes cosine similarity between CLIP visual features extracted from the masked region of the inpainted image and the corresponding ground-truth region.
After fine-grained assessment, we found that CLIP@mask does not always correlate with human preference in hallucination-sensitive scenarios.
As shown in Fig.~\ref{fig:supp-vlm-vs-clip}, CLIP@mask may assign high scores to semantically incorrect inpainting results or to cases that clearly exhibit hallucinated objects.
This limitation arises because CLIP-based metrics measure feature similarity rather than explicitly reasoning about object existence and semantic validity.
For this reason, the main paper adopts VLM-based hallucination judgment and validates its trend with human evaluation.

\begin{figure}[h]
\centering
\includegraphics[width=\linewidth]{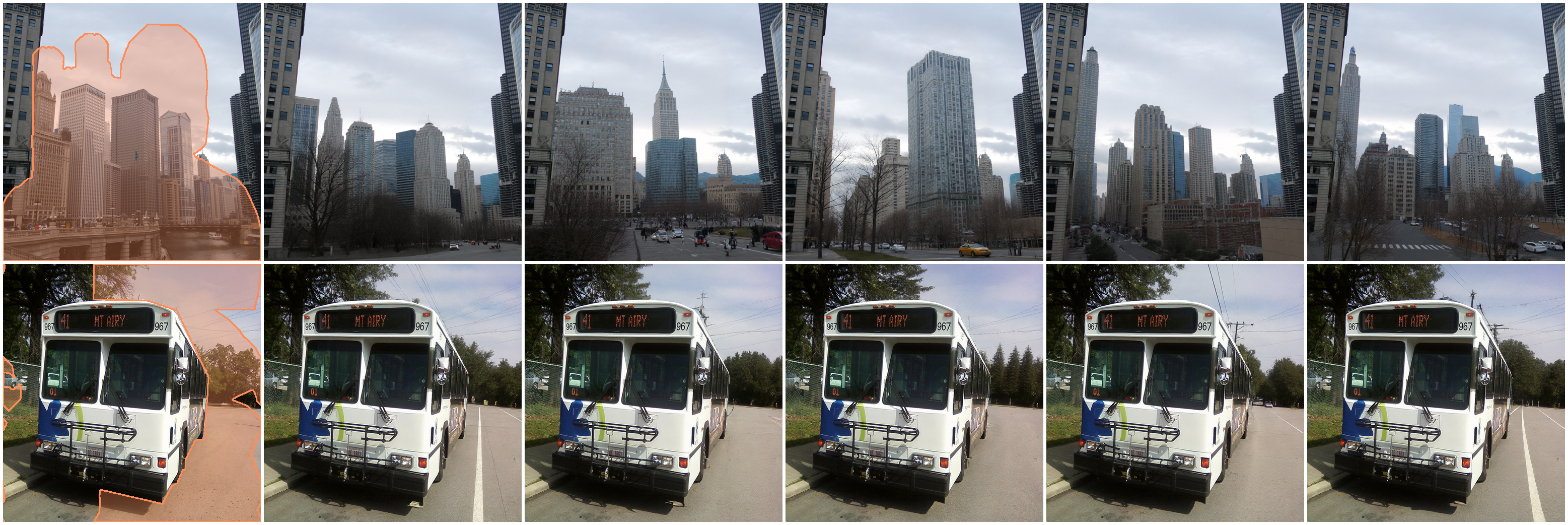}
\caption{\label{fig:diversity}
\textbf{Cross-seed diversity of ASUKA-II-FLUX.} The leftmost column shows the
input with the masked region highlighted; the remaining columns are completions
from five different seeds under the same image and mask.}
\end{figure}

\paragraph{Diversity}
As shown in Fig.~\ref{fig:diversity}, fixing the input image and mask and
varying only the sampling seed yields diverse completions, indicating that our method retains good generative diversity.

\begin{table}[h]
\centering
\footnotesize
\setlength{\tabcolsep}{3pt}
\caption{
  Quantitative comparison on MISATO@512. Each cell is the mean $\pm$ sample standard
  deviation over five seeds.
}
\label{tab:misato512_seedvar}
\begin{tabular*}{\linewidth}{@{\extracolsep{\fill}}lccccc}
\toprule
Model & LPIPS~$\downarrow$ & FID~$\downarrow$ & U-IDS~$\uparrow$ & P-IDS~$\uparrow$  & G@e~$\downarrow$  \\
\midrule
\midrule
OmniEraser
 & 0.2963 {\tiny $\pm$ 0.0141}
  & 35.71 {\tiny $\pm$ 3.03}
  & 0.1204 {\tiny $\pm$ 0.0140}
  & 0.0237 {\tiny $\pm$ 0.0038}
  & 190.93 {\tiny $\pm$ 20.68}\\
OmniPaint
& 0.1933 {\tiny $\pm$ 0.0074}
  & 14.90 {\tiny $\pm$ 0.66}
  & 0.3697 {\tiny $\pm$ 0.0092}
  & 0.1942 {\tiny $\pm$ 0.0140}
  & 84.81 {\tiny $\pm$ 3.28}\\
PixelHacker
& 0.1507 {\tiny $\pm$ 0.0002}
  & 12.67 {\tiny $\pm$ 0.07}
  & 0.3513 {\tiny $\pm$ 0.0038}
  & 0.1716 {\tiny $\pm$ 0.0095}  
  & 54.45 {\tiny $\pm$ 0.33}
\\
\midrule
ASUKA-II-FLUX
& \textbf{0.1399} {\tiny $\pm$ 0.0009}
  & \textbf{10.70} {\tiny $\pm$ 0.08}
  & \textbf{0.3862} {\tiny $\pm$ 0.0063}
  & \textbf{0.2403} {\tiny $\pm$ 0.0090}
  & \textbf{48.36} {\tiny $\pm$ 0.21}
\\
\bottomrule
\end{tabular*}
\end{table}

\paragraph{Statistical Analysis}
We evaluate all methods on MISATO@512 and each method is executed five times with seeds
$\{42, 1234, 2025, 7, 999\}$, and Table~\ref{tab:misato512_seedvar} reports the
mean $\pm$ sample standard deviation over those five runs. Our method is the most accurate of the four and is stable under reseeding. It achieves the best results with stable behavior: the standard deviation
across seeds small.

\begin{table}[h]
\centering
\footnotesize
\setlength{\tabcolsep}{6pt}
\caption{
  Inference cost on MISATO@512. Steps is the number of denoising steps of each
  method's default sampler. Time is the steady-state inference time for a single
  image on one H200, averaged over all shards and seeds.
}
\label{tab:misato512_cost}
\begin{tabular*}{\linewidth}{@{\extracolsep{\fill}}lcc}
\toprule
Model & Denoising steps & Time (s)~$\downarrow$ \\
\midrule
\midrule
OmniEraser          & 28 & 2.90 \\
OmniPaint           & 28 & 5.02 \\
\midrule
ASUKA-II-FLUX & 50 & 2.91 \\
\bottomrule
\end{tabular*}
\end{table}

\paragraph{Efficiency Analysis}
Table~\ref{tab:misato512_cost} reports the number of denoising steps and the per-image inference time of each method that based on FLUX. 
Our method performs a lightweight adaptation on top of a \emph{frozen} FLUX.1-Fill backbone, and therefore inherits the sampling schedule of the base model: we deliberately do not modify the number of denoising steps ($50$). 
Per denoising step our method takes $0.058$\,s,
whereas the FLUX-LoRA erasers need $0.104$\,s (OmniEraser) and $0.179$\,s (OmniPaint). 
Among the methods that adapt a frozen FLUX backbone, ours is the most efficient per step.

\begin{figure}[t]
\centering
\includegraphics[width=0.75\linewidth]{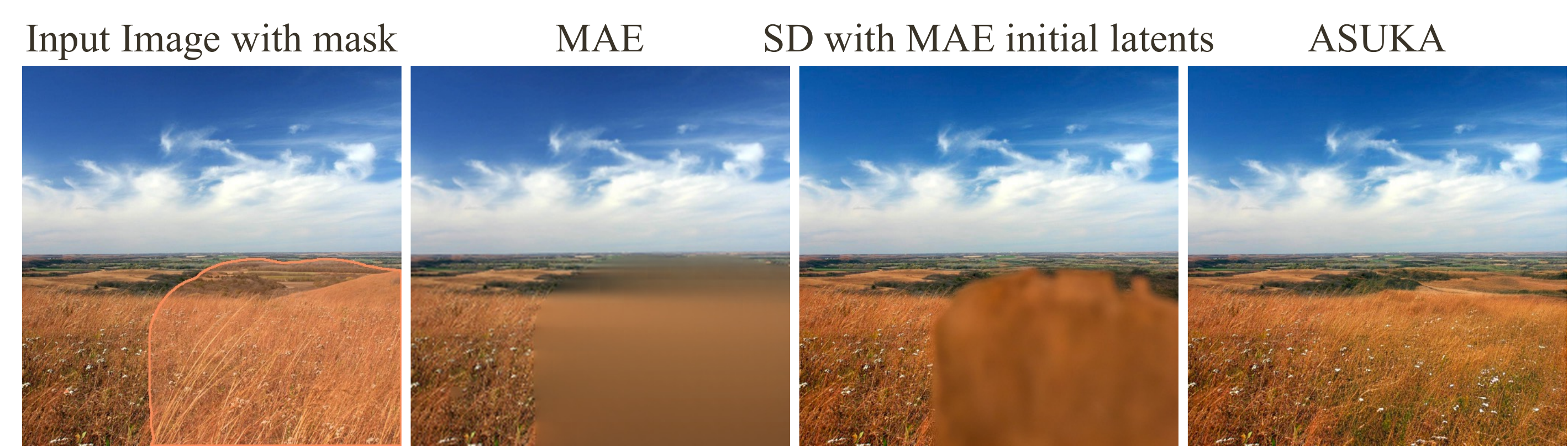}
\caption{Using the MAE prior for direct image-to-image translation through SD gives poor inpainting results. This motivates using MAE as a stabilizing condition rather than as the final output.}
\label{fig:supp-mae-to-image}
\end{figure}

\begin{figure}[t]
\centering
\includegraphics[width=0.4\linewidth]{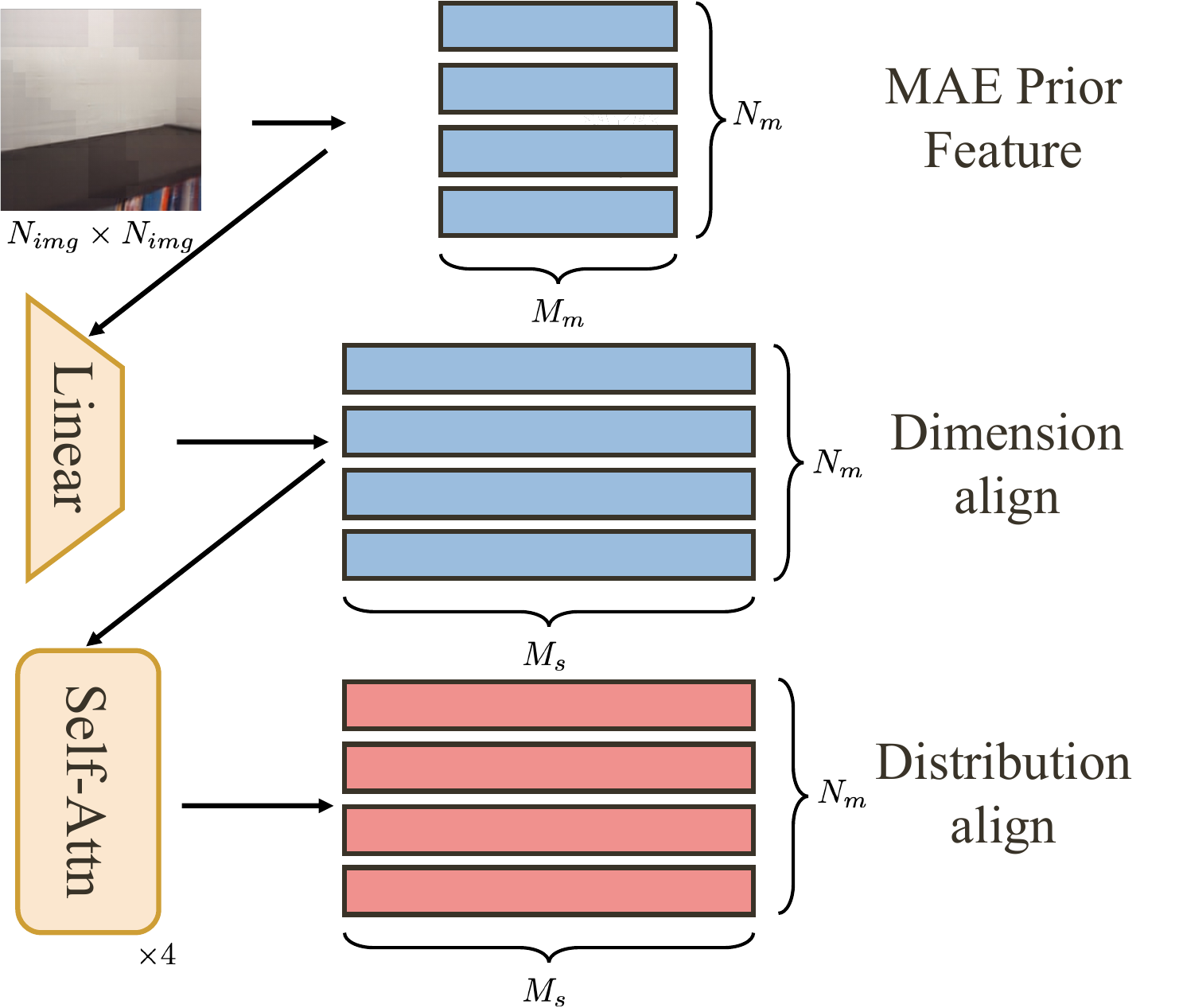}
\caption{The alignment module for MAE prior and generator contains a linear dimension alignment and a self-attention distribution alignment.}
\label{fig:supp-alignment}
\end{figure}

\begin{figure}[t]
\centering
\includegraphics[width=0.8\linewidth]{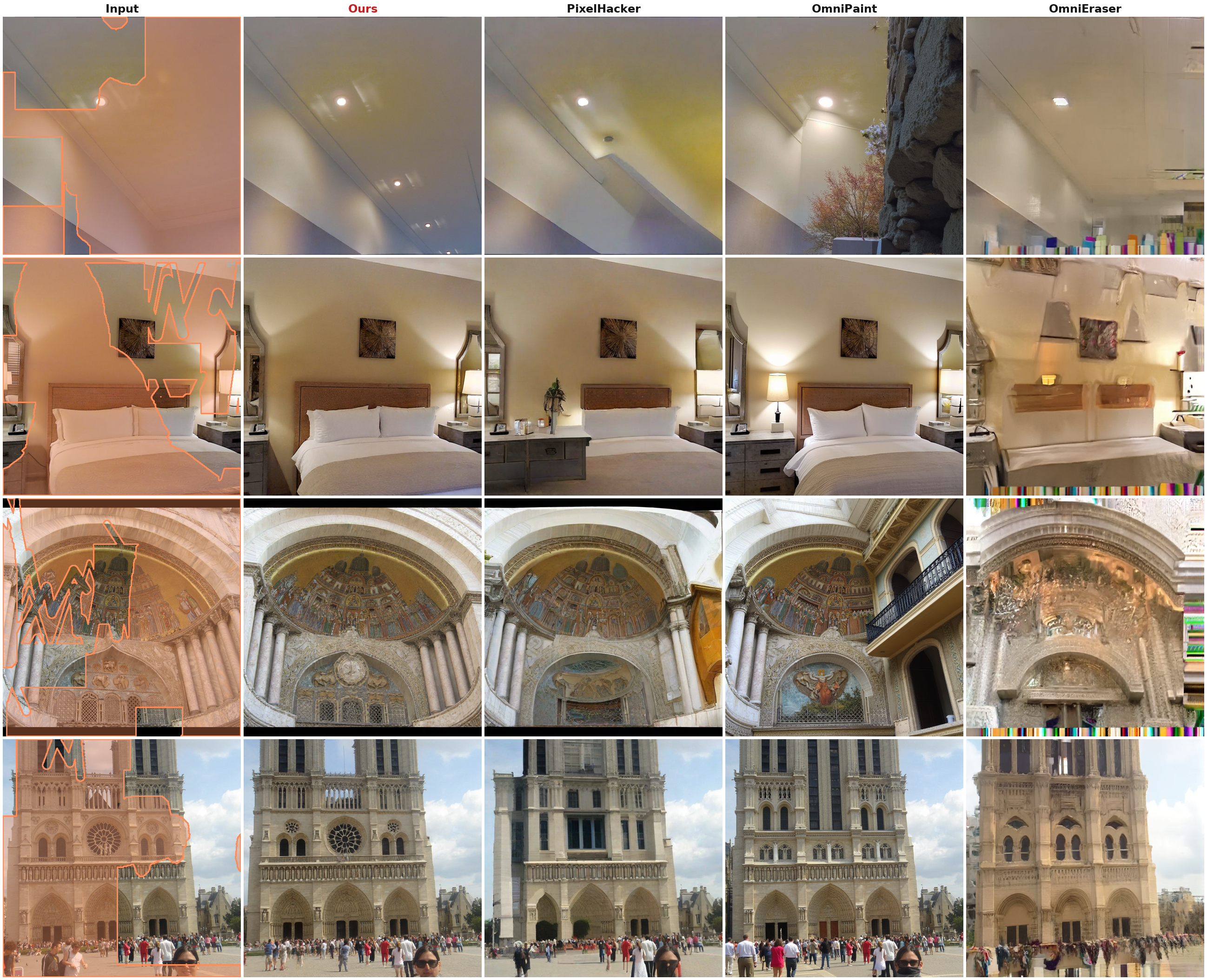}
\caption{\label{fig:comparison}
Qualitative comparison with recent SOTA inpainting models.}
\end{figure}

\begin{table}[t]
\centering
\begin{threeparttable}
\caption{\label{tab:user-study}User-study of top-1 ratio among all the inpainting results. 
}
\begin{tabular*}{\linewidth}{@{\extracolsep{\fill}}lcc}
\toprule
Model & UOM (\%) & CC(\%) 
\\
\midrule
\midrule
Co-Mod~\cite{zhao2021large} & 3.98  & 4.98  \\
MAT~\cite{li2022mat} & 7.40 & 3.20\\
LaMa~\cite{suvorov2022resolution} & 8.18 & 8.28\\
MAE-FAR~\cite{cao2022learning} & 4.88 & 5.60\\
\midrule
SD~\cite{Rombach_2022_CVPR} & 10.58 & 5.75\\
SD-text  & 7.70& 15.83 \\
SD-prompt  & 16.18 & 15.78 \\
SD-Repaint~\cite{lugmayr2022repaint} & 1.60 & 0.55\\
\midrule
ASUKA-I-SD & \textbf{39.43} & \textbf{40.05} \\
\bottomrule
\end{tabular*}
\end{threeparttable}
\end{table}%

\ifdefined\ASUKASupplementaryInput
\else
  \bibliographystyle{IEEEtran}
  \bibliography{main}
\fi

\ASUKAEndSupplementary

\fi

\end{document}